\documentclass[runningheads]{llncs}

\usepackage[year=2024]{eccv}

\usepackage{eccvabbrv}

\usepackage{graphicx}
\usepackage{booktabs}
\usepackage{chngcntr}

\usepackage[accsupp]{axessibility}  %

\usepackage[dvipsnames]{xcolor}
\usepackage{makecell}
\usepackage{booktabs}
\usepackage{siunitx}
\usepackage{xcolor}         %
\usepackage{wrapfig}
\usepackage{subcaption}
\usepackage{mwe}
\usepackage{graphbox} %
\usepackage{pgfplots} 
\usepackage{tabularx} %

\usepackage{titletoc} %

\usepackage{amsmath,amsfonts,bm}

\def\eqref#1{equation~\ref{#1}}

\def\1{\bm{1}}

\DeclareMathAlphabet{\mathsfit}{\encodingdefault}{\sfdefault}{m}{sl}
\SetMathAlphabet{\mathsfit}{bold}{\encodingdefault}{\sfdefault}{bx}{n}

\newif\ifcomments

\commentstrue

\newcommand{\newtext}[3]{\ifcomments {{#3}} \else {#3} \fi}

\newcommand{\New}[1]{\newtext{NEW}{magenta}{#1}}

\definecolor{nvgreen}{HTML}{76B900}
\definecolor{myblue}{HTML}{5364cc}
\newcommand{\ours}{\textsc{Latte3D}\xspace}

\newcommand{\largeDataset}{\textit{gpt-101k}\xspace}
\newcommand{\animalStyle}{\textit{animal-style}\xspace}
\newcommand{\animalReal}{\textit{animal-real}\xspace}

\newcommand{\renderer}{\mathbf{R}}
\newcommand{\modelF}{\mathcal{M}}
\newcommand{\geometryNet}{G}
\newcommand{\textureNet}{T}

\newcommand{\inputShape}{s}

\newcommand{\camera}{c}
\newcommand{\inputText}{p}

\newcommand{\outputShape}{o}
\newcommand{\blend}{\alpha}
\newcommand{\loss}{\mathcal{L}}
\newcommand{\lossInit}{\loss_{\textnormal{recon}}}

\newcommand{\lossTrain}{\loss_{\textnormal{train}}}

\newcommand{\lossReg}{\loss_{\textnormal{reg}}}

\newcommand{\lossText}{\loss_{\textnormal{SDS}}}

\usepackage{xspace}
\newcommand{\generateTime}{400ms\xspace}

\newcommand{\mappingNetEmbedding}{{\mathbf{v}}}
\newcommand{\latentTriplane}{{\mathbf{h}}}

\usepackage[pagebackref,breaklinks,colorlinks,citecolor=eccvblue]{hyperref}

\usepackage{orcidlink}
\usepackage{url}
\usepackage{multirow}
\usepackage{booktabs}
\usepackage{graphicx}
\usepackage{siunitx}
\usepackage{wrapfig}
\usepackage{bbm}
\usepackage{wrapfig}
\usepackage{tikz}
\usetikzlibrary{positioning}
\usepackage{verbatim}
\usepackage{tabularx}
\usepackage{adjustbox}

\begin{document}

\newcommand{\titleString}{\textcolor{myblue}{LATTE3D}: \textcolor{myblue}{L}arge-scale \textcolor{myblue}{A}mortized \textcolor{myblue}{T}ext-\textcolor{myblue}{T}o-\textcolor{myblue}{E}nhanced\textcolor{myblue}{3D} Synthesis\vspace{-0.0225\textheight}}
\title{\titleString}

\newcommand{\authorHspace}{\hspace{0.01\textwidth}}

\newcommand{\titlerunningString}{LATTE3D}
\titlerunning{\titlerunningString}

\newcommand{\authorString}{
Kevin Xie$^{*}$ \and
Jonathan Lorraine$^{*}$ \and
Tianshi Cao$^{*}$ \and
Jun Gao \and \newline
James Lucas \and
Antonio Torralba \and
Sanja Fidler \and
Xiaohui Zeng
}
\author{\authorString}

\newcommand{\authorrunningString}{K.~Xie et al.}
\authorrunning{\authorrunningString}

\newcommand{\instituteString}{\vspace{-0.01\textheight}NVIDIA}
\institute{\instituteString\footnotetext{*Authors contributed equally.}}

\maketitle

\begin{figure}[htbp]
    \vspace{-.0425\textheight}
    \makebox[\linewidth][c]{%
        \includegraphics[width=1.35\linewidth,trim={0.0cm 0 0 0},clip]{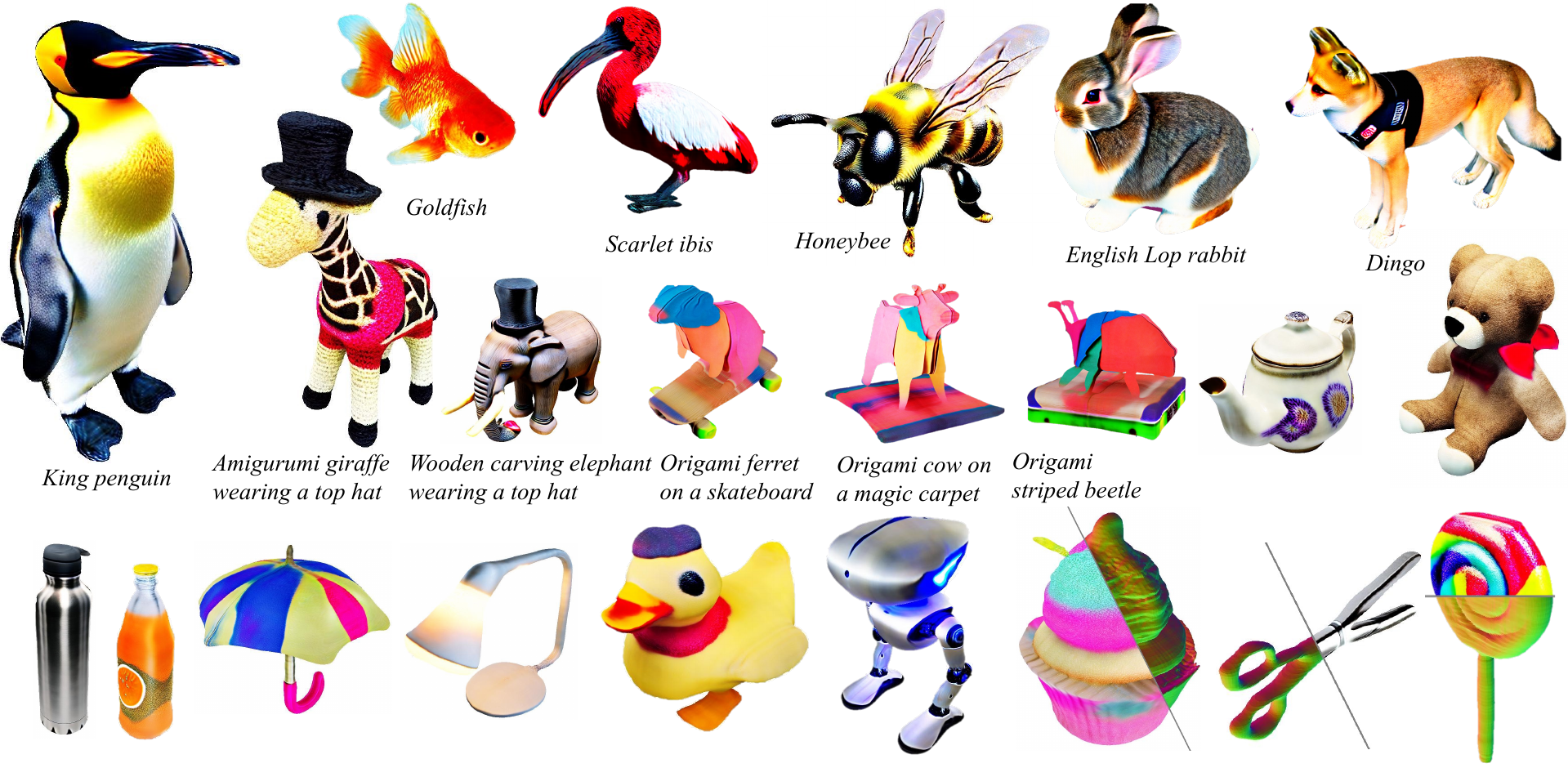}%
    }
    \vspace{-7.25mm}
    \captionof{figure}[Short caption]{\small\textbf{Samples generated in $\sim$\generateTime on a single A6000 GPU from text prompts. %
    } 
    Objects without prompt labels are generated by our text-to-3D model trained on $\sim\!100$k prompts, while labeled objects are generated by our 3D stylization model trained on $12$k prompts.
    See the \href{https://research.nvidia.com/labs/toronto-ai/LATTE3D/}{project website} for more.
    }
    \label{fig:teaser-top}
    \vspace{-.0575\textheight}
\end{figure}

\begin{abstract}
\noindent
Recent text-to-3D generation approaches produce impressive 3D results but require time-consuming optimization that can take up to an hour per prompt~\cite{poole2022dreamfusion,lin2023magic3d}. Amortized methods like ATT3D~\cite{lorraine2023att3d} optimize multiple prompts simultaneously to improve efficiency, enabling fast text-to-3D synthesis.
However, they cannot capture high-frequency geometry and texture details and struggle to scale to large prompt sets, so they generalize poorly. 
We introduce \ours, addressing these limitations to achieve fast, high-quality generation on a significantly larger prompt set. 
\New{Key to our method is 1) building a scalable architecture and 2) leveraging 3D data during optimization through 3D-aware diffusion priors, shape regularization, and model initialization to achieve robustness to diverse and complex training prompts. }
\ours amortizes both neural field and textured surface generation
to produce highly detailed textured meshes in a single forward pass. \ours generates 3D objects in \generateTime, and can be further enhanced with fast test-time optimization. %

\end{abstract}    
\newpage
\vspace{-5.5mm}
\section{Introduction}
\label{sec:intro}
\vspace{-1.5mm}

\vspace{-1.5mm}
Recent advances in text-to-3D synthesis via pre-trained image diffusion models mark significant progress in democratizing 3D content creation using natural language~\cite{poole2022dreamfusion,lin2023magic3d,wang2023prolificdreamer,fantasia,sjc,threestudio2023,metzer2022latent}. 
However, these methods often involve an expensive and time-consuming optimization that can take up to an hour to generate a single 3D object from a text prompt~\cite{poole2022dreamfusion,lin2023magic3d}.
To turbocharge these tools, we want techniques to generate various high-quality 3D objects in real-time, enabling rapid iteration on outputs and empowering user creativity in 3D content creation. We aim to achieve real-time text-to-3D synthesis for diverse text prompts and support fast test-time optimization when a further quality boost is desired.

\begin{wrapfigure}{r}{0.4\textwidth}
    \vspace{-7.0mm}
  \centering
  
    \centering
    \includegraphics[width=1\linewidth]{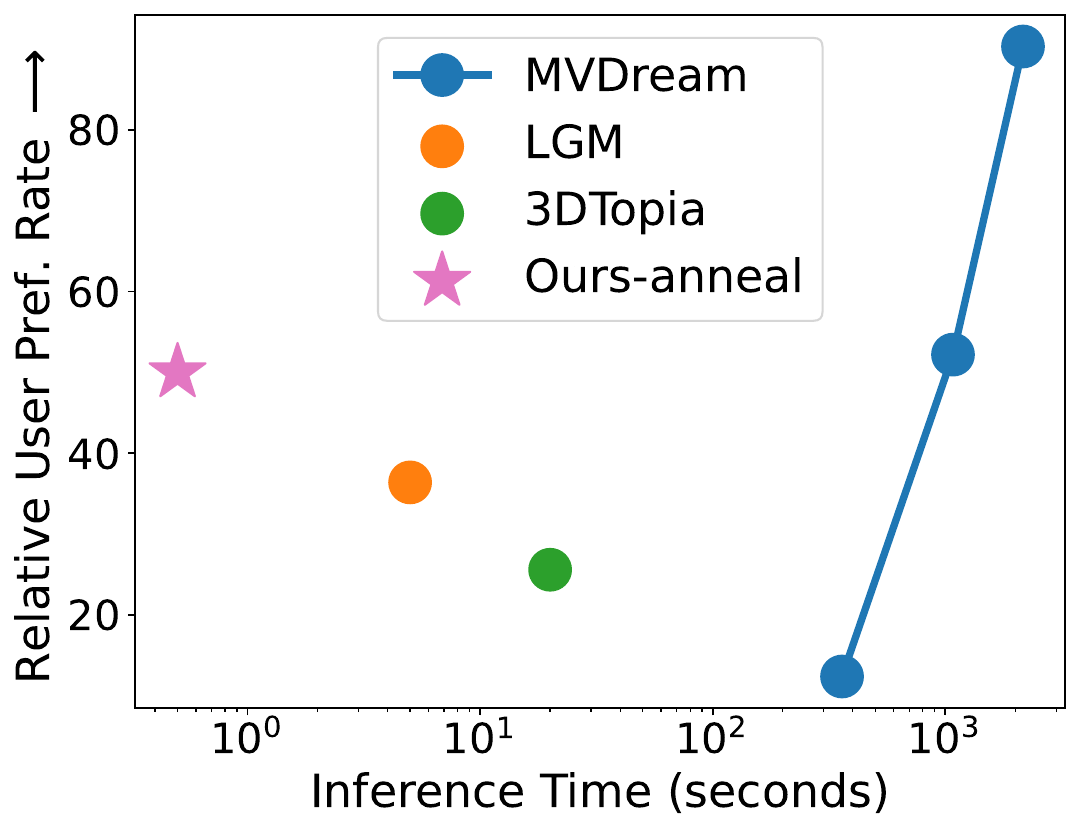}
  \vspace{-0.04\textheight}
  \caption{A quantitative comparison of SOTA text-to-3D methods on unseen prompts. %
    We plot different methods' user study preference rates compared to \ours.
    For MVDream, we report results with varying optimization times.
    }
  \label{fig:vs_baseline_inference}
  \vspace{-8.0mm}
\end{wrapfigure}

Seminal work~\cite{poole2022dreamfusion} proposed to optimize a neural field using the score distillation sampling (SDS) loss to generate 3D assets. It unlocked open-vocabulary 3D content creation but was (a) lacking high-frequency geometric and texture details, (b) expensive, and (c) prone to failing to generate the desired object.
Follow-ups introduced a surface-based fine-tuning stage~\cite{lin2023magic3d,wang2023prolificdreamer,fantasia} to generate high-frequency details. While this two-stage pipeline defines most current state-of-the-art text-to-3D approaches, it remains expensive and prone to prompt failure. 

ATT3D~\cite{lorraine2023att3d} was the first to make text-to-3D generation fast by simultaneously training a single model on a set of prompts in a process called \emph{amortized optimization}. %
Amortization reduces training time by optimizing a shared text-conditioned model on a set of prompts, enabling the model to generalize to new prompts at inference time.
Despite promising results on curated prompts, ATT3D -- and the underlying per-prompt methods -- remain prone to failing on general prompts. \New{ATT3D's simple architecture has limited capacity and lacks strong inductive biases for 3D generation, struggling to scale with dataset size and rendering resolution,
limiting the method to small scale (100s-1000s) prompt sets and low-fidelity textures.} 
Also, ATT3D only amortizes the first stage of 3D synthesis, producing a neural field representation and thus cannot generate high-frequency details.  

\New{To address these issues,} we introduce \ours, a {\bf L}arge-scale {\bf A}mortized {\bf T}ext-{\bf t}o-{\bf E}nhanced{\bf 3D} synthesis method that can produce high-quality 3D content in real-time. %
Our work makes the following technical contributions to boost the quality, robustness, scale, and speed of text-to-3D generative models: \vspace{-0.005\textheight}\New{
\begin{itemize}
    \item We propose a novel text-to-3D amortization architecture that can scale to orders of magnitude larger prompt sets.
    \item We leverage 3D data in training to improve quality and robustness, through 3D-aware diffusion priors, regularization loss, and weight initialization through pretraining with 3D reconstruction. 
    \item We amortize the surface-based refinement stage, greatly boosting quality.
\end{itemize}}

\vspace{-2.75mm}
\section{Related Work}
\vspace{-3.0mm}
Early work on 3D object synthesis was typically limited to generating objects from a given class~\cite{
wu2016learning,pointflow,gao2022get3d,achlioptas2018learning,zhou2021pvd,zeng2022lion,henzler2019platonicgan,lunz2020inverse,occnet,nash2020polygen,eg3d,tatarchenko2017octree}, 
\textit{e.g.} cars or chairs. Recent extensions use captioned 3D shapes to train text-to-3D models~\cite{jun2023shape,nichol2022point} generating shapes from diverse categories, but requiring 3D supervision, restricting them to synthetic datasets of limited size, diversity, and visual fidelity. 
 
The advent of differentiable rendering, both volumetric~\cite{mildenhall2020nerf} and surface-based~\cite{nvdiffrec}, opened the door to inverse image rendering~\cite{poole2022dreamfusion,gao2022get3d},
 unlocking the use of powerful text-to-image generative models~\cite{Rombach_2022_CVPR,dalle,imagen} in the 3D synthesis process. 
DreamFusion~\cite{poole2022dreamfusion} proposed the SDS loss to optimize a neural field using a text-to-image diffusion model to generate 3D assets. Follow-up work introduced a surface-based refinement stage~\cite{lin2023magic3d}, allowing the synthesis of high-frequency details. This two-stage optimization pipeline now defines most of the state-of-the-art text-to-3D approaches~\cite{wang2023prolificdreamer,fantasia,metzer2022latent,sjc}.
The recently released large-scale 3D dataset Objaverse-XL~\cite{objaverse,deitke2023objaversexl} has spurred researchers to explore the benefits of 3D supervision in the synthesis process. 
To date, the most successful approaches make text-conditioned image diffusion models 3D aware~\cite{liu2023zero1to3,shi2023MVDream,instant3d2023,shi2023zero123plus}
by fine-tuning 2D diffusion models using rendered multiview images. %
SDS from multiview diffusion models is then used to obtain 3D shapes ~\cite{shi2023zero123plus,shi2023MVDream,liu2023zero1to3}. %

However, these approaches require a lengthy, per-prompt optimization process, hindering applicability to real-world content creation.
To address this, two lines of work have been pursued. 
The first type of method uses a text-to-image-to-3D approach that generates images with a text-to-image diffusion model and trains an image-to-3D lifting network~\cite{instant3d2023,shi2023zero123plus,liu2023one, tang2024lgm}.
They offer improved speed but with limited quality details compared to optimization methods, but still take $5-20$ seconds per prompt~\cite{instant3d2023, tang2024lgm} for sampling from the image diffusion model.  
 
 In an alternate line of work, ATT3D~\cite{lorraine2023att3d} proposed an amortized framework that optimizes multiple prompts simultaneously with a unified model. Follow-up works expanded the network architecture to achieve better quality \cite{li2023instant3d, qian2024atom}.
\ours extends this line of work to larger scales by incorporating 3D knowledge to achieve prompt robustness during training, among other design improvements like better architecture. 
 Amortized text-to-3D offers an attractive quality vs. speed trade-off as it uses a single feed-forward architecture without requiring sampling of diffusion models - see Fig.~\ref{fig:vs_baseline_inference}.

\vspace{-4mm}
\paragraph{Concurrent Works}
Concurrently, AToM~\cite{qian2024atom} also amortizes two-stage training with a triplane representation, but they use DeepFloyd~\cite{deepfloyd} and do not scale beyond the smaller sets of thousands of prompts from ATT3D.
In contrast, we scale to the order of $\num{100000}$ prompts, using 3D-aware techniques such as MVDream.
Another concurrent work of ET3D~\cite{Chen2023ARXIV} achieves fast generation by training a GAN model, but it is only trained on a small-scale compositional dataset of $\num{5000}$ prompts, and 
only shows compositional generalization results like ATT3D and AToM. Furthermore, it does not guarantee view consistent 3D outputs as it uses a 2D neural upsampler.
Lastly, LGM~\cite{tang2024lgm} concurrently works in the text-to-image-to-3D direction that generates Gaussian splats in $5$ seconds, representing the latest state-of-art. We experimentally compare our method against this method and demonstrate competitive performance.

\vspace{-2mm}
\section{Methodology}
\label{sec:methodology}
\vspace{-1mm}

An established pipeline for high-quality text-to-3D generation consists of two stages, each performing per-prompt optimization using the SDS loss with a text-to-image diffusion model~\cite{lin2023magic3d,wang2023prolificdreamer,fantasia}. %
Stage-1 optimizes a volumetric representation, which is typically a neural radiance field. A coarse resolution is often used to speed up rendering.
Meshes afford real-time rendering
but are hard to optimize from scratch with image supervision alone. 
Stage-2 uses the output of the neural field to initialize a signed distance field (SDF) and a texture field, from which a mesh can be derived using differentiable isosurfacing~\cite{dmtet,shen2023flexicubes}. This surface-based representation is then optimized via differentiable rasterization~\cite{nvdiffrec}, which is fast even at $\num{1024}$ rendered image resolution -- enabling supervision from higher-resolution text-to-image diffusion models.

ATT3D~\cite{lorraine2023att3d} amortizes optimization of a neural field over a set of prompts with a hypernetwork mapping a text prompt to a neural field, which is trained with the SDS loss over a prompt set, referred to as \emph{seen prompts}. %
\ours introduces a new architecture that amortizes both stages of the generation process, aiming to produce high-quality textured meshes in real-time. %
We scale the approach to a magnitudes larger and more diverse set of prompts by leveraging 3D knowledge in the amortization loop. 

Our approach is illustrated in Fig.~\ref{fig:pipeline}.
We initialize \ours with a reconstruction pretraining step to stabilize training (Sec.~\ref{sec:pretraining}). 
The architecture consists of two networks, one for geometry and the other for texture (Sec.~\ref{sec:method_model_arch}). 
Amortized model training with diffusion priors is done through a two-stage pipeline consisting of a volume-based stage-1 and a surface-based stage-2 (Sec.~\ref{sec:method:stage_1}).
To reconcile reconstruction pretraining, which uses 3D geometry as input, with the task of text-to-3D generation, we anneal the network's input during stage-1 to gradually fully replace the dependence on 3D input with a single dummy input.
During inference (Sec.~\ref{sec:method_inference}), our model generates a 3D textured mesh from just a text prompt in \generateTime and allows an optional lightweight test-time refinement to enhance the quality of geometry and textures (Sec.~\ref{sec:method_test_time_optim}).

\vspace{-0.015\textheight}
\subsection{Pretraining to reconstruct shapes.} \label{sec:pretraining}
\begin{figure}[t]
    \centering
    \includegraphics[width=0.9\linewidth]{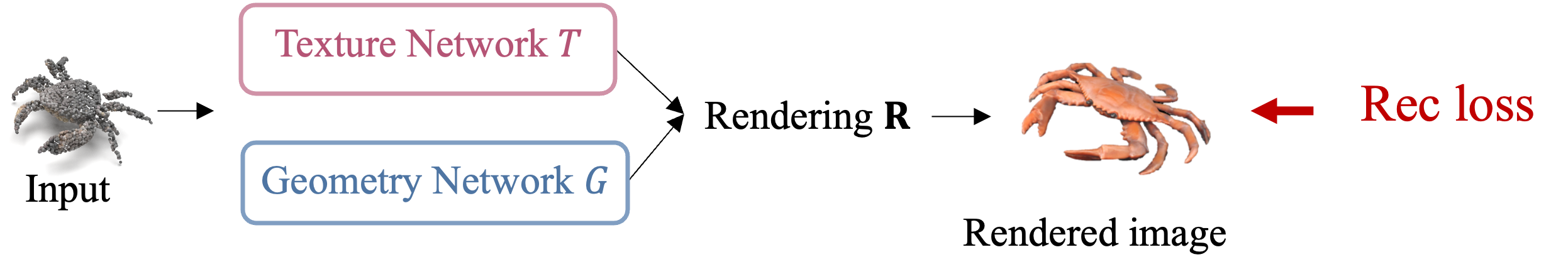}
    \vspace{-2mm}
    \caption{
        We overview our reconstruction pretraining here, which we use to achieve our shape initialization to improve prompt robustness.
    }
    \label{fig:rec}
\end{figure}

\vspace{-0.005\textheight}
Although SDS loss can synthesize text-aligned 3D features from scratch, empirically, it is found to be a less robust source of supervision due to high variance. A good initialization could stabilize the optimization process. 
We find that pretraining the model first to be able to encode-decode 3D shapes makes it easier to optimize with amortized SDS subsequently.
We initialize our model $\modelF$ by pretraining it on a dataset of 3D assets using image reconstruction losses, similar to the reconstruction stage used in~\cite{gupta20233dgen}, and we show the pipeline in Fig.~\ref{fig:rec}. $\modelF$ takes as input a sampled point cloud and outputs a predicted 3D shape.
We then render multiple views of the predicted 3D shape $\outputShape$ to compare with the input shape $\inputShape$ renderings. We use an $\ell_2$ loss on the rendered opacity and RGB image:  \vspace{-1.5mm}
\begin{equation}
\resizebox{0.9\linewidth}{!}{$
\lossInit(\outputShape, \inputShape, \camera) = ||  \renderer_{\text{opacity}}(\outputShape, \camera) - \renderer_{\text{opacity}}(\inputShape,\camera)||_2 + ||\renderer_{\text{RGB}}(\outputShape, \camera) - \renderer_{\text{RGB}}(\inputShape,\camera)||_2
$} \vspace{-1.5mm}
\label{eq:loss_pretrain}
\end{equation}
We denote the renderer $\renderer$, the opacity and RGB rendering with $\renderer_{\text{opacity}}$ and $\renderer_{\text{RGB}}$, respectively, and the camera with $\camera$, randomly sampled on a sphere.

\subsection{Model Architecture}\label{sec:method_model_arch}
Our model $\modelF$ consists of two networks, $\geometryNet$ and $\textureNet$, for predicting geometry and texture, respectively. After pretraining, we introduce text-conditioning by adding cross-attention layers. We use CLIP~\cite{clip} embeddings for encoding the text prompt $\inputText$. In stage-1 training, we tie the geometry and texture networks' weights (except for separate MLP decoder heads), effectively training a single input encoder. In stage-2, we freeze the geometry network $\geometryNet$ and refine the texture network $\textureNet$ with an additional trainable upsampling layer. 
The final model $\modelF$ used in inference is the frozen $\geometryNet$ from stage-1 and a refined $\textureNet$ from stage-2. %

\begin{figure*}[t]
    \centering
    \includegraphics[width=0.97\linewidth]{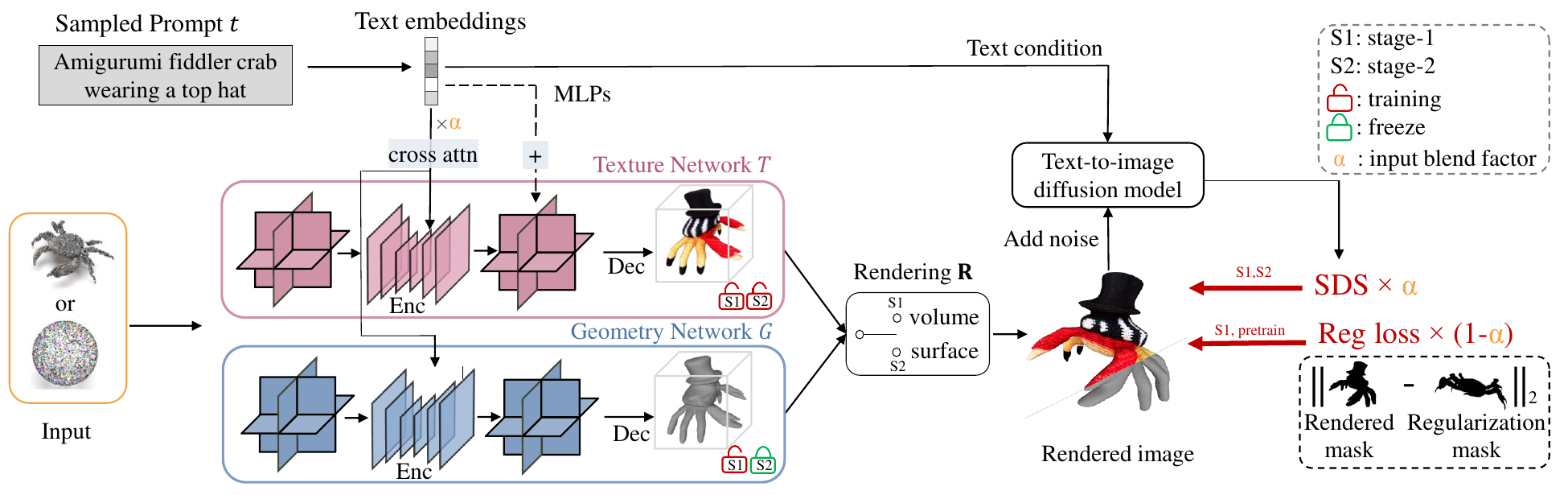}
    \vspace{-2mm}
    \caption{ %
    \small \ours consists of two networks: a texture network $T$ and geometry network $G$. When amortizing the first stage, the encoders of both networks share the same set of weights. The training objective includes an SDS gradient from a 3D-aware image prior and a regularization loss that compares the rendered predicted shape's mask with the rendered masks of 3D assets in a library. When amortizing surface-based refinement in stage-2, we freeze the geometry network $G$ and update the texture network $T$. %
    }
    \label{fig:pipeline} 
    \vspace{-3.2mm}
\end{figure*}

\vspace{-3mm}
\paragraph{The geometry network} $\geometryNet$ %
consists of a U-Net encoder on a triplane representation~\cite{eg3d,gao2022get3d}, followed by a decoder containing another triplane U-Net and a neural volumetric density field which predicts the output shape from triplane features.
\New{Specifically, a point cloud is fed through a PointNet, and the features are converted to a triplane representation via average scattering based on geometric projection to each plane. During training, the point cloud comes from the 3D data, and in inference, the point cloud is replaced by a dummy input of a sphere point cloud.}
The triplane representation is processed through the encoder and decoder U-Nets. 
The text embedding is fed into every residual U-Net block in the encoder via cross-attention. 
For any point in 3D space, we extract triplane features by projecting into each plane and bilinearly interpolating the feature map. 
An MLP then predicts the density value at that point.

\vspace{-3mm}
\paragraph{The texture network} $\textureNet$ %
and geometry network $\geometryNet$ share the same encoder and decoder U-Net architecture in pretraining and stage-1. 
We employ another non-shared MLP neural texture field to predict RGB values at query points.
In stage-2, we upsample the triplane features produced by the encoder to a higher resolution to gain extra capacity to generate high-frequency texture details. Specifically, we first bilinearly upsample the latent triplane from $128$ resolution to $480$ and add a residual MLP which maps the text-embedding to the residual of the triplane feature, mimicking ATT3D's \emph{mapping (hyper)network}. The decoder then takes this upsampled triplane as input.
Details are in the Appendix.

\vspace{-3mm}
\paragraph{Rendering.}
To train the model, we render the generated 3D object into 2D images with different methods in stage-1 and 2.
In stage-1, the output 3D geometry is a triplane-based neural field, on which we use volume rendering to get images of $256$ resolution using the formulation from VolSDF~\cite{yariv2021volume}, allowing us to parameterize the density via an approximate signed distance field whose zero-level set defines the object geometry. This choice allows a simple conversion to surface-based rendering. For the geometry network, we observe more stable training %
with volume rendering with SDS compared to surface rendering.
In stage-2, we extract the isosurface from the density field with Marching Cubes~\cite{lorensen1998marching}  %
and render images via rasterization~\cite{nvdiffrast}. Meshes can be rendered at a $\num{1024}$ image resolution in real-time, helping capture finer details for learning textures.

\begin{figure*}[t]
    \vspace{-0.015\textheight}
    \centering
    \includegraphics[width=\linewidth, trim={0 5cm 0 0}, clip]{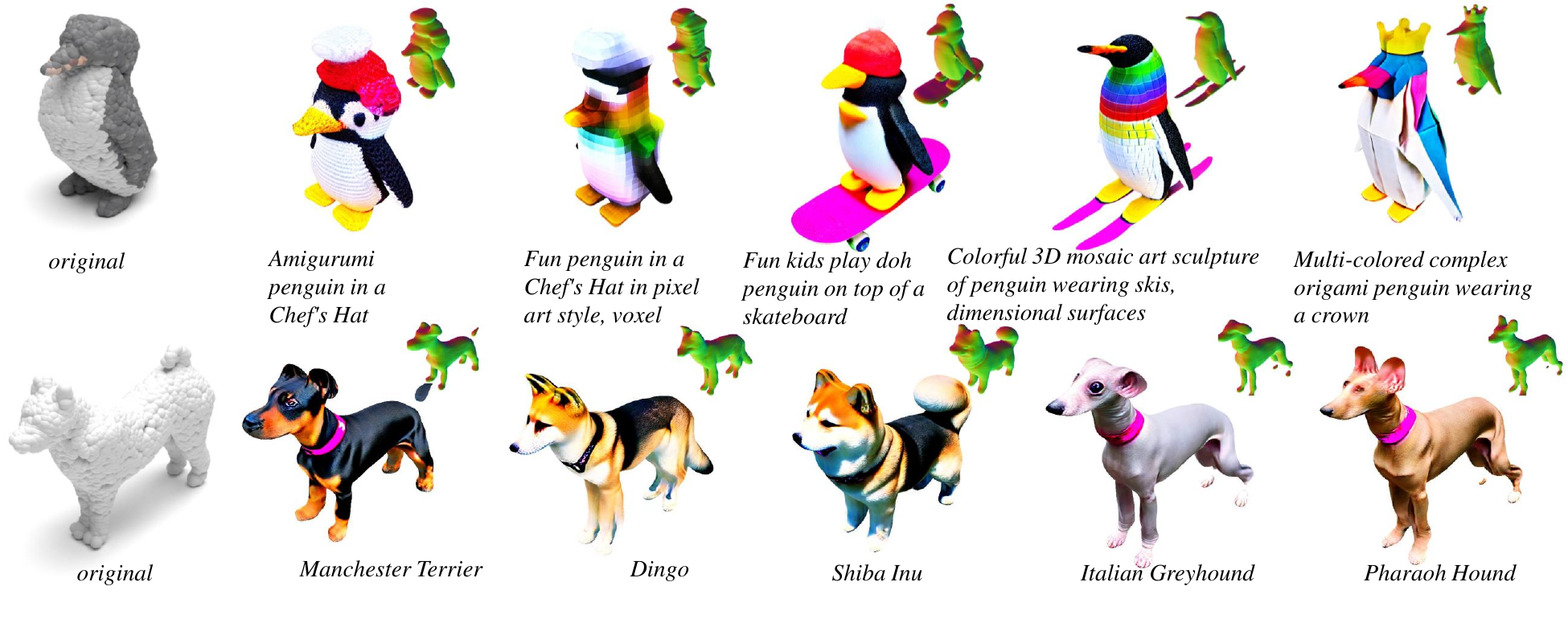}
    \vspace{-8mm}
    \caption{\textbf{Stylization application}: Our model learns to generate diverse stylizations of the same shapes. Left column shows the original shape to be stylized. }
    \label{fig:example_prompt}
    \vspace{-0.015\textheight}
    \end{figure*}
\newcolumntype{Y}{>{\centering\arraybackslash}X} %

\subsection{Amortized Learning} 
\label{sec:method_using_3d_data}

Amortized learning distills knowledge from image diffusion priors into our model.
We perform amortized training on volumetric geometry training (stage-1) and surface texture training (stage-2) sequentially.
We incorporate 3D information into our training process (a) implicitly via 3D-aware 2D SDS losses and (b) explicitly with 
regularization.

\vspace{-3mm}
\paragraph{Curating prompt sets.}
\label{sec:amortization_data}
We create large prompt sets for training using rule-based text generation or ChatGPT~\cite{OpenAI2023GPT4TR}. For the former, we start with the categories names from the captions of Objaverse~\cite{luo2023scalable}, and design rules, like ``object A in style B is doing C'' with a set of styles and activities.

Alternatively, we input the captions into ChatGPT and ask it for detailed and diverse prompts describing a similar 3D object. 
Fig.~\ref{fig:pipeline} shows that when generating \emph{``Amigurumi fiddler crab wearing a top hat''}, we guide shape synthesis towards resembling a crab by retrieving a related 3D shape used in a regularization loss.
Other details of dataset construction are in Sec.~\ref{exp:dataset}. 

\vspace{-4mm}
\subsubsection{Amortizing Stage-1 Generation}
\label{sec:method:stage_1}
In this stage, we train $\modelF$ with our datasets. We use the SDS loss with a 3D-aware 2D prior and regularization to the paired 3D shape, which we now explain in detail:

\vspace{-3mm}
\paragraph{3D-aware SDS loss.} During stage-1 training
we leverage a 3D aware diffusion prior, which provides a stronger, multiview consistent supervisory signal. Specifically, we use the model from MVDream~\cite{shi2023MVDream}, which was trained on four rendered views of objects from the Objaverse dataset~\cite{objaverse} by modifying Stable Diffusion~\cite{Rombach_2022_CVPR} to generate multiview consistent images. 

\vspace{-3mm}
\paragraph{Regularization loss.} 
While 3D-aware SDS loss with MVDream offers a strong multiview prior, it can still fail to correctly capture the full 3D geometry for some prompts since MVDream only supervises four views and can create enlarged geometry or floaters in the shape. Some failures are shown on the right of Fig.~\ref{fig:blend_compare_stylized}. %

Combining regularization using 3D shapes with the SDS loss helps training and improves geometry.   
As shown in Fig.~\ref{fig:pipeline}, for each training prompt, we regularize the output shape $\outputShape$ by comparing its rendered mask to the mask of a shape $\inputShape$ retrieved from the 3D dataset with the prompt: 
\vspace{-1mm} \begin{equation} 
    \lossReg(\outputShape, \inputShape, \camera) = ||\renderer_{\textnormal{opacity}}(\outputShape, \camera) - \renderer_{\textnormal{opacity}}(\inputShape, \camera)||_2,
\end{equation}
where $\renderer_{\textnormal{opacity}}$ is the opacity mask (volume-rendered density).
We balance the regularization $\lossReg$ and the SDS loss $\lossText$ by simply blending them using a convex combination with weight $\blend$. 
Our final training loss is:
\vspace{-1mm} \begin{equation}
\label{eq:blending}
\lossTrain = (1 - \blend) \lossText + \blend \lossReg. 
\end{equation}
During training, we render five views for each generated shape. We average across four views to compute $\lossText$, and one is used to compute $\lossReg$.

\vspace{-3mm}
\paragraph{Input point cloud annealing} \New{To bridge the gap between the real point cloud inputs in training and dummy point cloud in inference, we gradually anneal the input point clouds towards dummy point clouds during stage-1 training. Specifically, we randomly replace a subset of points in the input point cloud with points from the dummy point cloud using a probability that increases linearly from $0$ to $1$ in the last $5000$ training iterations. With point cloud annealing, the model gets better results in inference when the dummy point cloud is used. } %

\vspace{-1mm}
\subsubsection{Amortizing Stage-2 Generation}
\label{sec:method:stage_2}
During stage-2 training, we freeze the network for our geometry and only tune the texture since training both leads to instabilities. We use depth-conditional ControlNet~\cite{zhang2023adding} for SDS guidance, allowing a higher, $512$ resolution supervision. Depth conditioning encourages the SDS loss to guide the texture to align with the geometry from the geometry network, thereby improving the 3D texture consistency.

\vspace{-2mm}
\subsection{Inference} 
\label{sec:method_inference}
\vspace{-1mm}
\New{During inference, our model inputs the user's text prompt and a dummy point cloud.}
Our $\modelF=(\geometryNet,\textureNet)$ outputs a textured mesh, where $\textureNet$ is the final texture network refined in stage-2, while $\geometryNet$ is trained in stage-1 only and then frozen.
Predicting one shape and rendering the image only takes us \generateTime on an A6000 GPU in inference, allowing the generation of user results at interactive speeds.
A single A6000 can generate $4$ samples simultaneously.
Once the user selects a desired sample, textured meshes can be exported from \ours. We use a UV mapping tool to UV parameterize the mesh output by $\geometryNet$ and use $\textureNet$ to predict the albedo colors of each texel in a UV-mapped texture image. The resulting mesh and texture image is compatible with standard rendering software.

\vspace{-2mm}
\subsection{Test Time Optimization} 
\label{sec:method_test_time_optim}
\vspace{-1mm}
Our method supports test time optimization if the user wants to boost the quality of a particular prompt. This can be useful if the user's prompt deviates significantly from the seen prompts. 
As in our stage-2 training, we freeze the geometry and finetune the texture network $\textureNet$ with depth-conditional SDS guidance. Our test time optimization is significantly faster to converge than MVDream~\cite{shi2023MVDream} or other optimization-based text-to-3D methods~\cite{lin2023magic3d,poole2022dreamfusion,wang2023prolificdreamer} since we optimize our amortized network instead of a randomly initialized network. 

\vspace{-2mm}
\subsection{3D Stylization} 
\label{sec:stylization}
\vspace{-1mm}
\ours can also be trained and deployed as a 3D stylization method, allowing users to cheaply create variations from existing 3D assets. During training, we skip the input point cloud annealing step to maintain point cloud reliance and amortize over a dataset where each 3D shape corresponds to many style prompts. Our blend of regularization and SDS guidance drives the model to produce shapes structurally similar to the original shapes yet semantically aligned with the style prompt. Amortization significantly reduces the computational cost per shape-prompt pair compared to optimizing each pair separately. 
During inference, point clouds from the training set can be combined with novel prompts as input to the model to produce variations of the original shape.
\begin{figure}[t]
    \vspace{-0.025\textheight} 
    \centering

    \includegraphics[width=0.95\linewidth]{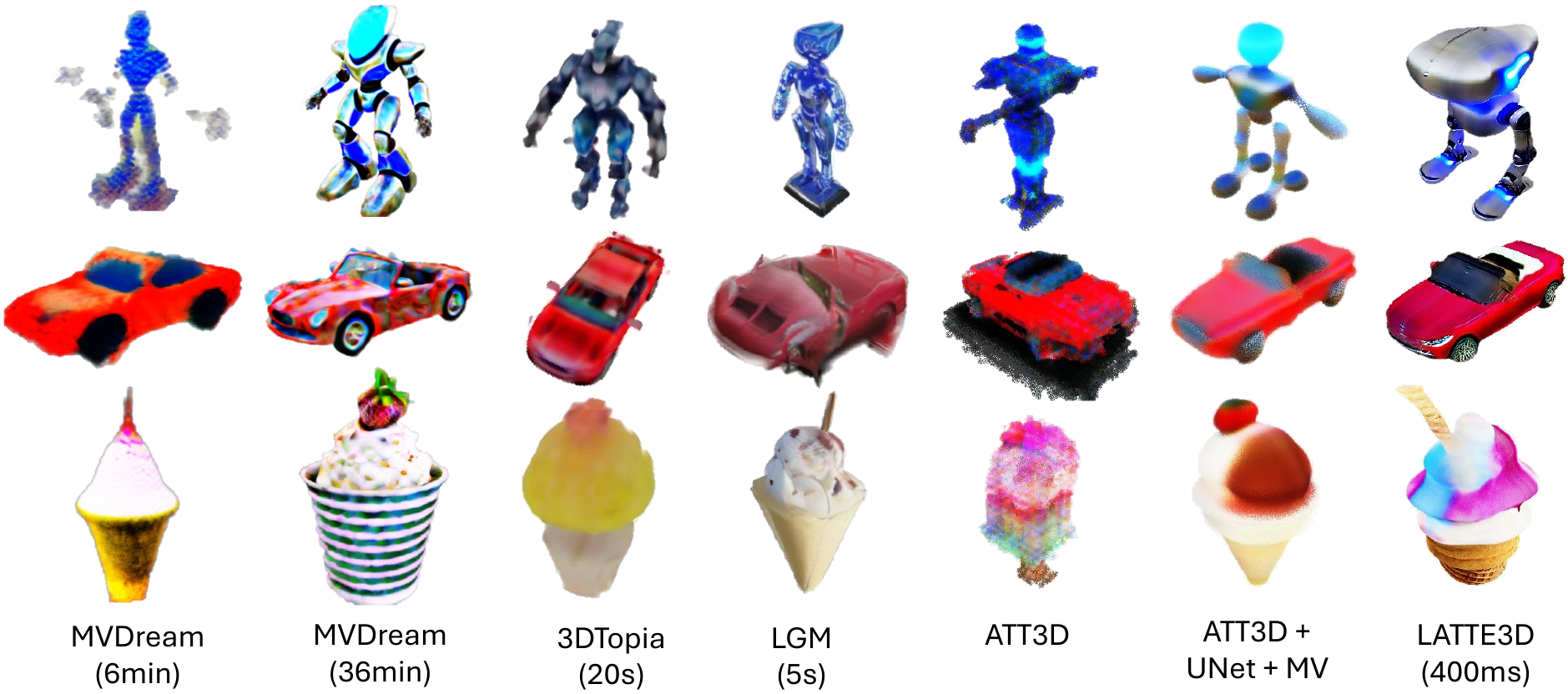} 
    \vspace{-0.01\textheight} 
    \caption{ \footnotesize{ %
        Qualitative comparison between \ours and ATT3D~\cite{lorraine2023att3d}, MVDream~\cite{shi2023MVDream}, Instant3D/3DTopia~\cite{instant3d2023} and LGM~\cite{tang2024lgm}.
        The top row is a training prompt: \textit{``a futuristic robot with shiny metallic armor and glowing blue accents..''}, and the last two rows are unseen prompts: \textit{``a red convertible car with the top down'', ``an ice cream sundae''}. }}

    \label{fig:vs_baselines}

    \vspace{-0.025\textheight}
\end{figure}

\section{Experiments}
We first describe our experimental setup, including datasets, metrics, and baselines for comparison (Sec.~\ref{sec:exp_setup}). We then present our quantitative and qualitative results (Sec.~\ref{sec:exp_main_exp}), demonstrating the performance and generalization ability of \ours (Sec.~\ref{sec:capability}), followed by design choices ablations (Sec.~\ref{sec:exp_ablation}). 

\vspace{-1mm}
\subsection{Experimental Setup}
\label{sec:exp_setup}
We now discuss dataset construction, model pretraining, and evaluation metrics.

\vspace{-2.5mm}
\subsubsection{Datasets}\label{exp:dataset}
\vspace{-1mm}
\begin{wraptable}{r}{0.5\textwidth}  
    \vspace{-0.4cm}
    \centering \footnotesize 
    \caption{\footnotesize Comparing the sizes of amortization training sets to prior works. }
    \scalebox{0.85}{
    \begin{tabular}{c c c} 
       \textbf{name} & \textbf{\#prompts} & \textbf{Source} \\ 
       \midrule
       Animal2400 \cite{lorraine2023att3d, qian2024atom} & \num{2400} & Rule based \\ 
       DF415 \cite{qian2024atom} & \num{415}  & DreamFusion\cite{poole2022dreamfusion} \\
       \animalStyle (Ours) & \num{12000} & Rule based \\
       \largeDataset (Ours) & \num{101608} & ChatGPT \\
    \end{tabular} }

    \label{tab:scale_of_data}
    \vspace{-1.0cm}
\end{wraptable}

\paragraph{Training Set} To investigate scaling amortized optimization beyond what has been done in prior works -- see Tab.~\ref{tab:scale_of_data} --  we construct a new dataset \largeDataset that consists of 101k text prompts and 34k shapes. 
For the shapes, we use the ``lvis'' subset from Objaverse \cite{objaverse} with 50k human-verified shapes. We filter out poor shapes, \eg flat images and scans, and retain 34k shapes for our dataset. We use ChatGPT to augment each object's caption to gather $3$ prompts per object. 
\paragraph{Evaluation Sets} We use two prompt sets for evaluation. To evaluate the benefit of amortized optimization in our method, we evaluate our model on \largeDataset. However, due to the computational cost of running the baselines, we randomly select a $50$ prompt subset in \largeDataset to form the \textbf{seen} prompt set for benchmarking. Furthermore, to gauge the generalization abilities of our amortized model, we construct an \textbf{unseen} prompt set by filtering a subset of $67$ DreamFusion\cite{poole2022dreamfusion} prompts that are close in distribution to lvis categories.

\subsubsection{Baselines and Evaluation Schemes}
\paragraph{Baselines.} We compare \ours with the baselines: ATT3D~\cite{lorraine2023att3d}, MVDream~\cite{shi2023MVDream}, Instant3D~\cite{instant3d2023} and LGM~\cite{tang2024lgm}. 
We re-implement ATT3D~\cite{lorraine2023att3d} \New{with a hypernetwork.} 
We use the open-source threestudio~\cite{threestudio2023} implementation of MVDream. For MVDream, we train each 3D model for $\num{10000}$ iterations, with a batch size of $8$ on a single A100 GPU with provided default settings. For Instant3D, since the source code has not been released, we use the re-implementation from 3DTopia\footnote{\url{https://github.com/3DTopia/3DTopia}}. For LGM, we use the official open-source implementation.

\paragraph{Evaluation metrics.} We quantitatively compare our model with baselines by evaluating the generated 3D content fidelity and its consistency to the text prompt for the optimization cost. We use three metrics to evaluate fidelity. 

\noindent
\textbf{Render-FID:}
We compute the FID~\cite{heusel2017gans} between the renderings of generated 3D content and a set of images sampled from Stable Diffusion~\cite{Rombach_2022_CVPR} with the same text prompts as input. This metric measures how well the generated shapes align with those from the 2D prior in visual quality. 

\noindent
\textbf{CLIP Score:} 
We compute the average CLIP scores between the text prompt and each rendered image to indicate how well the generated 3D shape aligns with the input text prompt. %

\noindent
\textbf{User Study:} We evaluate the overall 3D geometry and texture through Amazon Mechanical Turk user studies. For each baseline, we present videos of rendered 3D shapes generated by our method and a baseline side-by-side, with corresponding text prompts, and ask users for their preference. We present each comparison to three users and average the preference across users and prompts to obtain the average user preference for each baseline relative to \ours.

\noindent
\textbf{Timing:} 
We measure our \textbf{optimization cost} by GPU time per prompt to gauge the total compute cost on each prompt. We divide the total wall-clock time for pretraining, stage-1, and stage-2 by the prompt set's size, then multiply by the number of GPUs used. %
We measure \textbf{inference time} for Instant3D, LGM, and \ours by the time the model takes from in-taking the text input to outputting the final triplane features before rendering. 
For MVDream~\cite{shi2023MVDream}, we measure the average time to optimize a single prompt on an A100 GPU.

\begin{table}[t]

    \caption{\small  Quantitative metrics and average user preference ($\%$) of baselines over \ours trained on \largeDataset using \textbf{seen} and \textbf{unseen} prompts. We also report test-time optimization, which takes $10$ min. 
    }
  
    \centering
    \scalebox{0.9}{
    \setlength{\tabcolsep}{1pt} 
    \begin{tabular}{l|c|ccc c cc c }
        \toprule
        \multirow{2}{*}{Model} & Time & \multicolumn{2}{c}{Render-FID $\downarrow$} & \multicolumn{2}{c}{CLIP-Score $\uparrow$} & \multicolumn{2}{c}{Preference \% $\uparrow$}  \\
        &  & seen & unseen(df) & seen & unseen(df)  & seen & unseen(df) \\
        \midrule
        \ours & \generateTime & 180.58 & 190.00 & 0.2608 & 0.2605 & - & - \\
        \bottomrule
        \ours-opt  & $10$min & 171.69 & 178.37 & 0.2800 & 0.3000 & - & -\\
        \bottomrule 
        MVDream  & $6$min & 203.53 & 199.23 & 0.2240 & 0.2698 & 2.5 & 12.4  \\ %
        MVDream & $18$min & 191.77 & 188.84 & 0.2440 & 0.2950 & 26.5 & 52.2 \\ %
        MVDream & $36$min & 158.11 & 143.44 & 0.2830 & 0.3180 & 48.5 & 89.3\\ %
        \midrule
        3DTopia & 20s & 217.29 & 222.76 & 0.1837 & 0.2171 & 20.8 & 25.6\\
        LGM & 5s & 178.67 & 166.40 & 0.2840 & 0.3063 & 26.2 & 36.4 \\
        \bottomrule 
    \end{tabular} 
    
    }
    
    \label{tab:abs_scale}

    \vspace{-2mm}
\end{table}

\subsection{Experimental Results}
\label{sec:exp_main_exp}

\label{tab:vs_baseline_user}

We observe robust generalization of \ours by training on the \largeDataset dataset to \textbf{unseen} prompts (df, from DreamFusion~\cite{poole2022dreamfusion}), as demonstrated by qualitative (Fig.~\ref{fig:vs_baselines}) and quantitative (Fig.~\ref{fig:vs_baseline_inference}, Tab.~\ref{tab:abs_scale}) results.
We highlight the relative user preference of \ours versus different baselines over inference time in Fig.~\ref{fig:vs_baseline_inference}. \ours produce results of competitive user preference to SOTA baselines at a uniquely fast inference time.

\begin{wrapfigure}{b}{0.55\textwidth}  
    \vspace{-0.01\textheight}
    \centering 
    \centering
    \includegraphics[width=\linewidth]{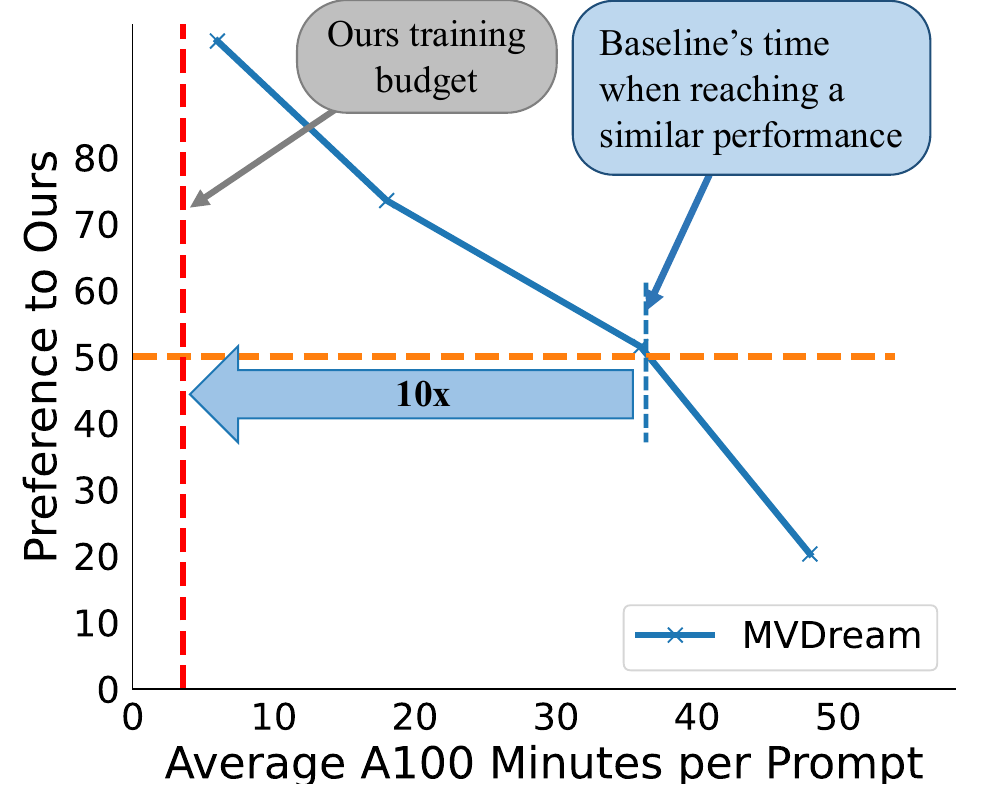}
    \vspace{-0.03\textheight}
    \caption{ \small 
    Results of user study showing the average preference of MVDream at different amounts of optimization time to \ours \largeDataset on seen prompts.
    }
    \label{fig:lvis_vs_baseline_speed}

    \vspace{-0.04\textheight}
\end{wrapfigure}  

From the quantitative comparisons in Tab.~\ref{tab:abs_scale}, our performance on both \textbf{seen} and \textbf{unseen} prompts is competitive with baselines while our inference time is at least one order of magnitude faster. We note that 3DTopia\cite{tang2024lgm} and LGM\cite{tang2024lgm}, which reconstruct 3D representations from four input views, are advantaged under image-based metrics (FID and CLIP score) as they tend to produce 3D inconsistent results that are not obvious from still-images. Interested readers can refer to rendered videos of all methods in the supplementary materials. 

\vspace{-0.02\textheight}
\subsubsection{Total optimization cost} A benefit of amortized optimization over many prompts is that our total optimization cost on \largeDataset is much lower than optimizing MVDream per-prompt. Fig.~\ref{fig:lvis_vs_baseline_speed} illustrates that to achieve user preference on par with \ours, MVDream would require $36$ GPU minutes per prompt, whereas \ours spent only $215$ GPU seconds per prompt, representing a $10\times$ reduction in optimization cost.

\vspace{-3mm}
\subsection{Application}\label{sec:capability}
We illustrate \ours's capacity for enhanced quality through test time optimization and how to adapt \ours for stylizing 3D content.

\begin{wrapfigure}{l}{0.45\textwidth}  
\vspace{-0.05\textheight}
    \centering
        \setlength{\tabcolsep}{0.1pt} 
    
  \begin{tabular}{ccc} 
    \includegraphics[width=0.33\linewidth, trim={0 0 0 3cm},clip]{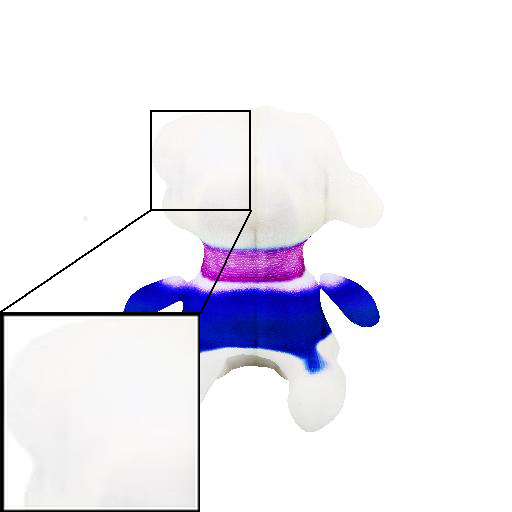} &
    \includegraphics[width=0.33\linewidth, trim={0 0 0 3cm},clip]{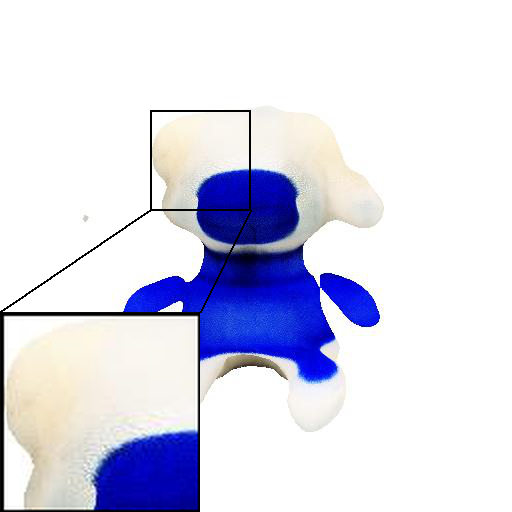} &
    \includegraphics[width=0.33\linewidth, trim={0 0 0 3cm},clip]{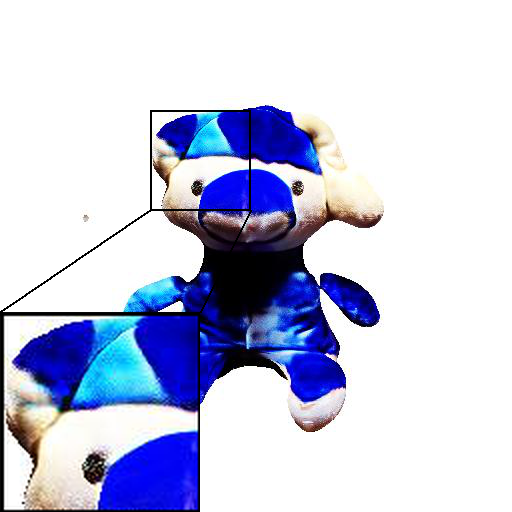} \\
    \includegraphics[width=0.33\linewidth, trim={0 0 0 3cm},clip]{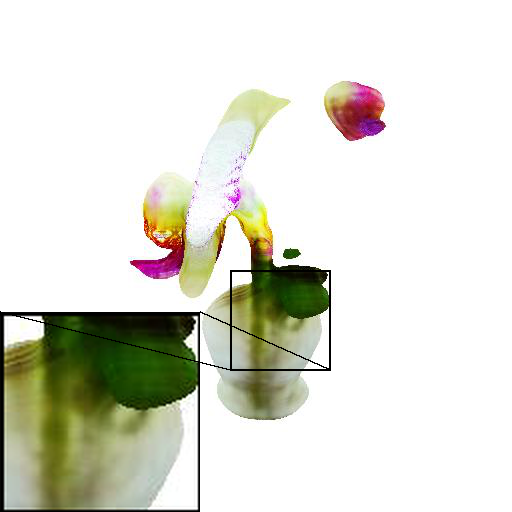} &
      \includegraphics[width=0.33\linewidth, trim={0 0 0 3cm},clip]{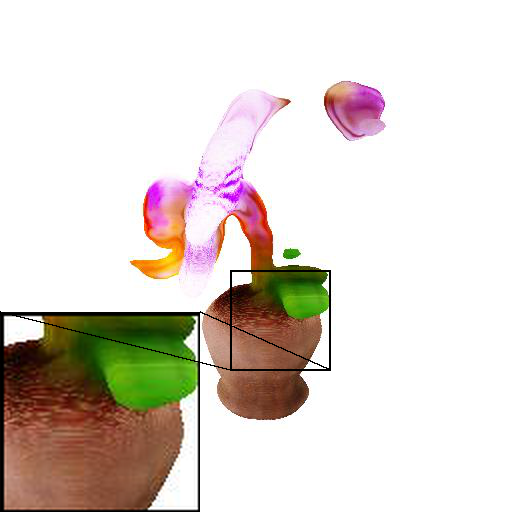} &
      \includegraphics[width=0.33\linewidth, trim={0 0 0 3cm},clip]{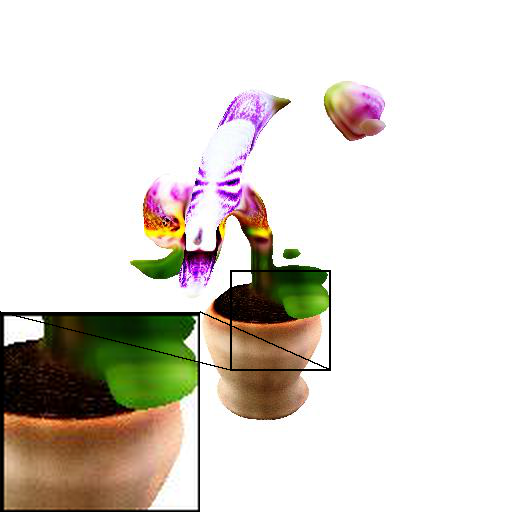} \\ 
      
  \generateTime &  $1$ min  & $10$ min \\ 
  \end{tabular}
    \vspace{-0.0125\textheight}
    \caption{\small \textbf{Test time optimization on seen and unseen prompt}. The model is trained on the \largeDataset dataset. First column: without test time optimization, \generateTime. Other columns: test time optimization with the given time budget denoted. 
    Top: seen prompt \textit{``..animal with a blue and white floral pattern..''} Bottom: unseen prompt \textit{``orchid in clay pot''.}
    }
    \label{fig:post_optimization}

    \vspace{-0.05\textheight}
\end{wrapfigure}

\vspace{-3mm}
\subsubsection{Per Prompt Test-time Optimization} 
We show experimentally that \ours quickly adapts to new prompts during test time to improve quality.
In Fig.~\ref{fig:post_optimization}, we qualitatively compare \ours with up to $600$ steps.
Each optimization iteration takes an average of $0.98$ seconds on an A100 GPU, giving us a total compute budget of less than $10$ GPU minutes per prompt.
Even with test-time optimization, our method is still an order of magnitude faster than per-prompt optimization methods such as MVDream, as in Tab.~\ref{tab:abs_scale}.
Test-time optimization is particularly beneficial on unseen prompts, where FID drops by $11.6$ for unseen vs. $8.8$ for seen, and the CLIP score gains $0.04$ for unseen vs. $0.02$ for seen. 

\vspace{-3mm}
\subsubsection{Stylizing 3D content}
\begin{figure}
    \vspace{-.25cm}
    \centering
    \includegraphics[width=0.95\linewidth]{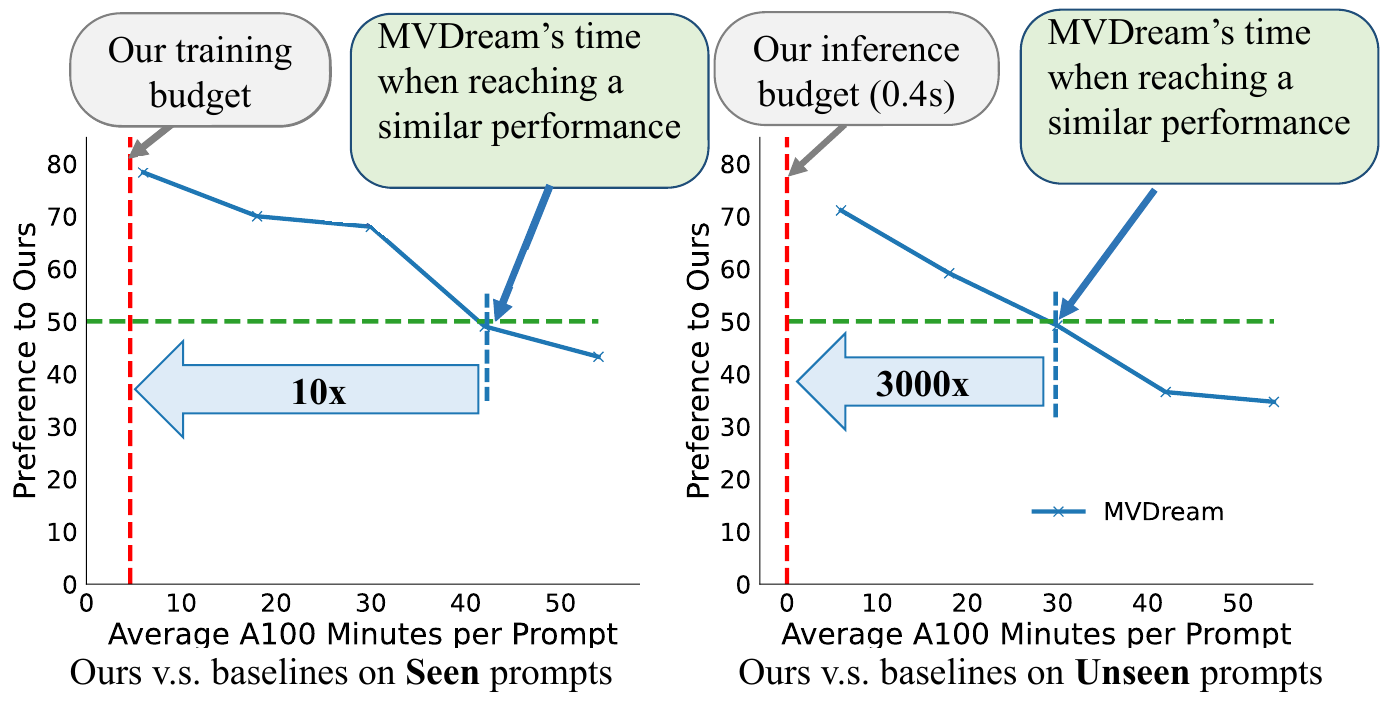}
    \vspace{-0.25cm}
    \caption{
    Results of user study showing the average preference rate for MVDream at different amounts of optimization time to \ours stylization results on \animalStyle.
    }
    \label{fig:animal_vs_baseline_speed}
    \vspace{-.5cm}
\end{figure}

We illustrate how \ours architecture offers us the flexibility to adapt it for targeted 3D content stylization as described in Sec.~\ref{sec:stylization}. 
To test this, we manually curate a dataset of 100 animal shapes from Objaverse as a basis and augment the species name of each animal shape name with combinations of activities (\eg ``riding a skateboard'') and styles (\eg ``voxel style'') to create our 12000 prompts \animalStyle dataset.
We train a \ours model (stage-1 and stage-2) on \animalStyle without the point cloud annealing stage. 
The resulting model adapts to both the text prompt and the point cloud inputs.
In Fig.~\ref{fig:example_prompt}, each row contains the outputs of \ours given different prompts while fixing the point cloud input, demonstrating the effectiveness of \ours for text-controlled stylization.
In Fig.~\ref{fig:animal_vs_baseline_speed}, we conduct a user study comparing \ours's outputs on \animalStyle to MVDream and find that \ours is competitive with MVDream with $\num{4000}$ steps of optimization, representing a $10\times$ reduction in optimization cost (compared to running MVDream for each prompt). \ours on \animalStyle also generalizes to held-out combinations of animal activities and styles with little quality drop, as indicated by its similar relative preference to MVDream. 
Therefore, \ours enables users to cheaply create variations of their 3D assets via amortized optimization and even explore novel prompt combinations at test time.
Additional quantitative and qualitative results can be found in the supplementary materials.

\subsection{Ablation Studies}
\label{sec:exp_ablation}
We now analyze the value of each component we have introduced in \ours, with more results in the Appendix. %

\begin{table}[t]
    \centering
    \caption{\small Ablation of components in stage-1 training. Trained on \largeDataset data and evaluated on seen and unseen prompts. Preference indicate average user preference of baseline over \ours. \ours is better than all ablated settings in quantitative metrics and is preferred by users on the unseen prompt set.
    }
    \label{tab:aba_method}
    \scalebox{0.85}{
    \setlength{\tabcolsep}{2pt} 
    \begin{tabular}{l|cccc|ccccc}
      \toprule
       \multirow{2}{*}{Exp} & \multirow{2}{*}{MV} & \multirow{2}{*}{Unet} & \multirow{2}{*}{Pretrain} & \multirow{2}{*}{Reg} & \multicolumn{4}{c}{Seen} & Unseen \\ \cmidrule{6-9}
        & & & & & Mask-FID$\downarrow$ & Render-FID$\downarrow$ & Clip-Score$\uparrow$& \multicolumn{2}{c}{Preference } \\
      \midrule

       ATT3D && & & & 274.44 & 275.01 & 0.2091 & 28.3 & 24.2 \\
       +MV &$\checkmark$& & & & 243.24  &  236.73 & 0.1826 & 34.8 & 26.9 \\
       +MV+UNet &$\checkmark$& $\checkmark$ & & & 184.80 & 203.33 & 0.2214 & 48.0 & 45.6  \\
       +MV+UNet+PT &$\checkmark$& $\checkmark$ & $\checkmark$ &   & 189.09 & 189.51 & 0.2191 & 51.8 & 47.7 \\
        \hline
       \ours (S1) &$\checkmark$& $\checkmark$ & $\checkmark$ & $\checkmark$& {\bf{176.44}} & {\bf{186.84}} & {\bf{0.2379}} & - & -\\ 
      \bottomrule
    \end{tabular}
    }
    \vspace{0.01\textheight}

\vspace{-0.5cm}
\end{table}

\vspace{-3mm}
\paragraph{Stage-1 ablation.} 
Tab.~\ref{tab:aba_method} summarizes our quantitative comparison of the following ablations: \textit{ATT3D} baseline using a hypernet, \textit{ATT3D+MV} uses MVDream as diffusion guidance. \textit{ATT3D+UNet+MV} replaces the hypernet with our model architecture, but is initialized from scratch and given a dummy point cloud as input. \textit{ATT3D+UNet+MV+PT} is initialized from the reconstruction pretrained (PT) model.
We introduce the Mask-FID metric to quantitatively assess geometry adherence to the shape dataset by computing the FID between the rendered binary masks of the generated shapes and those of the 3D dataset.

We find that each of the components we introduce improves performance. 
Comparing ATT3D with ATT3D+MV, we see benefits from using the MVDream guidance as it mitigates geometric artifacts like Janus faces. 
Comparing ATT3D+MV with ATT3D+MV+UNet we see a performance boost from our architecture. 
Further, adding pretraining (ATT3D+UNet+MV+PT) helps, especially in recovering fine geometric and texture details. Finally, adding shape regularization (full \ours) makes the generated geometry adhere better to the 3D data, as evidenced by the drop in Mask-FID. An additional user study is shown in Tab.~\ref{tab:aba_method}. We will show the convergence speed in the Appendix.

\begin{figure}
    \vspace{-0.2cm}
    \centering
    
    \begin{minipage}[c]{0.45\textwidth}
            \centering
    \includegraphics[width=\linewidth]{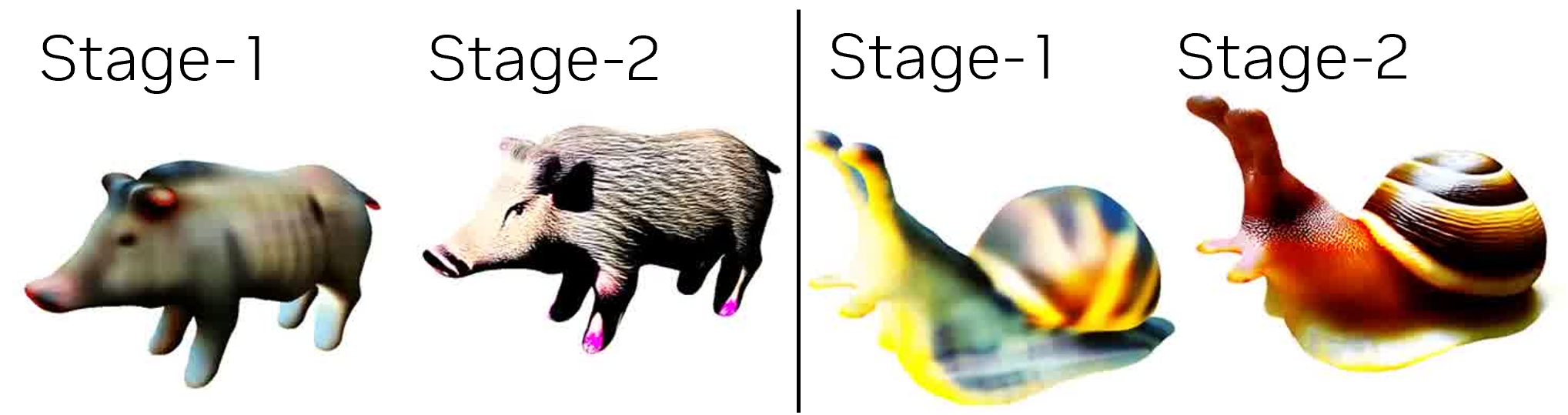}
    
    \caption{Comparison of stage-1 and stage-2 results. Stage-2 refinement significantly improves the texture quality. 
    }
    \label{fig:abaltion_stage}
    \end{minipage}
    \hspace{0.05\textwidth}
      \begin{minipage}[c]{0.45\textwidth}

    \small 
    \centering 
    \setlength{\tabcolsep}{0.2pt}
    \setlength\extrarowheight{-10pt}
    \tiny{{\textit{``fun sheep wearing skis in pixel art style, voxel''}}}\\\hphantom{a}\\
    \small
\begin{tabular}{cccc}
   
    0.3 & 0.2 & 0.1 & 0 \\
    \includegraphics[width=0.24\linewidth]{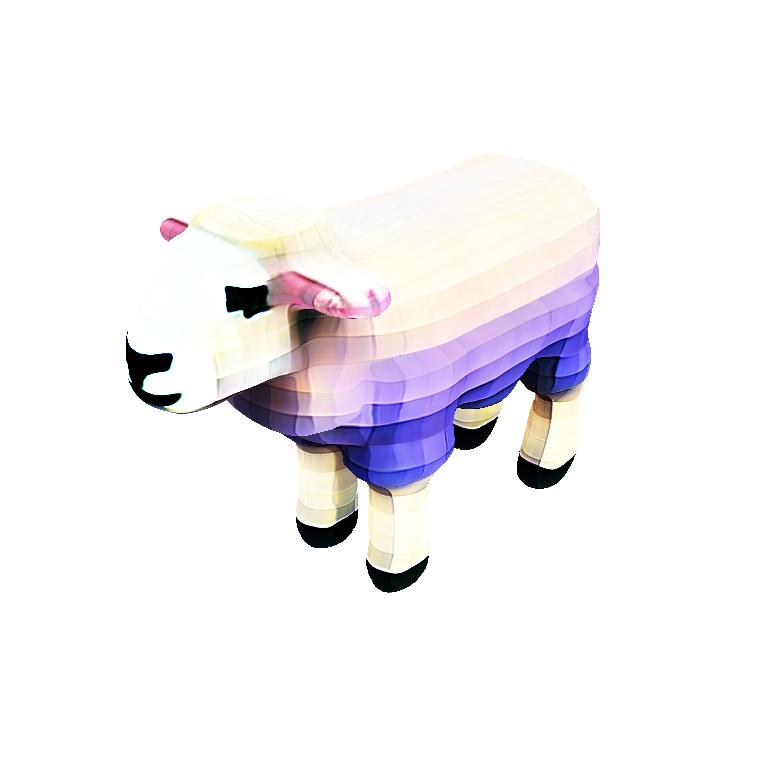} &
    \includegraphics[width=0.24\linewidth]{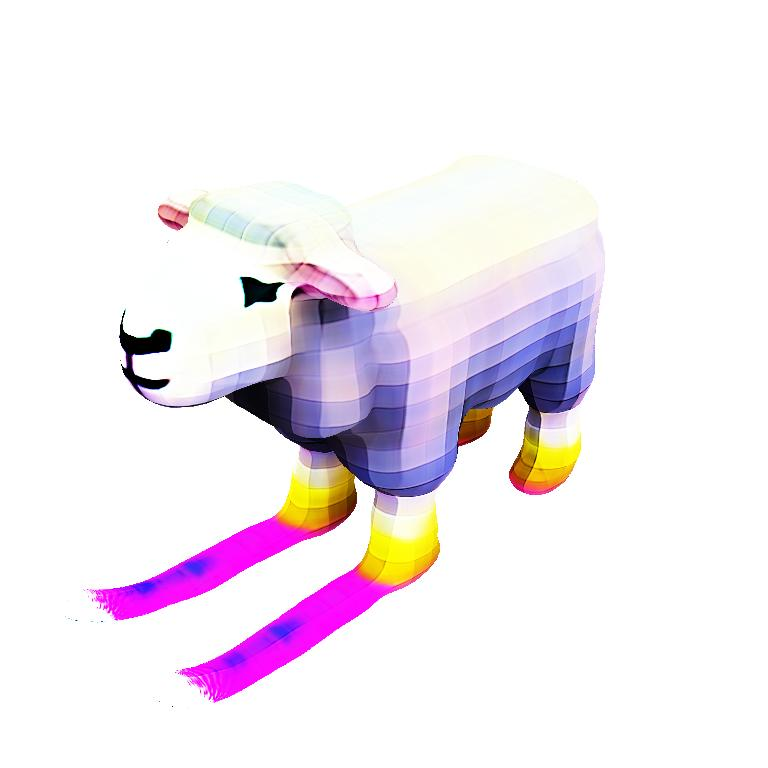} &
    \includegraphics[width=0.24\linewidth]{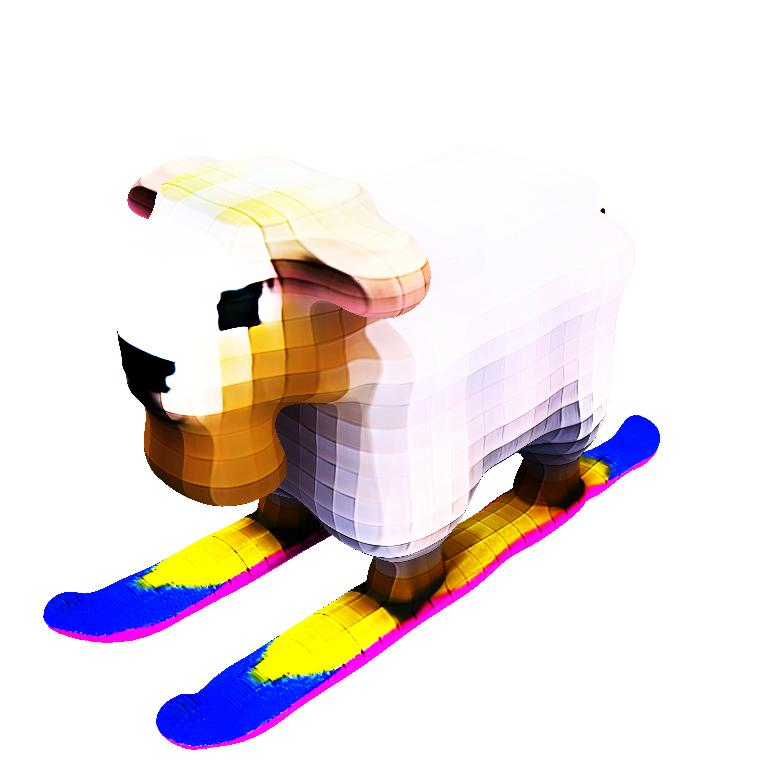} &
    \includegraphics[width=0.24\linewidth]{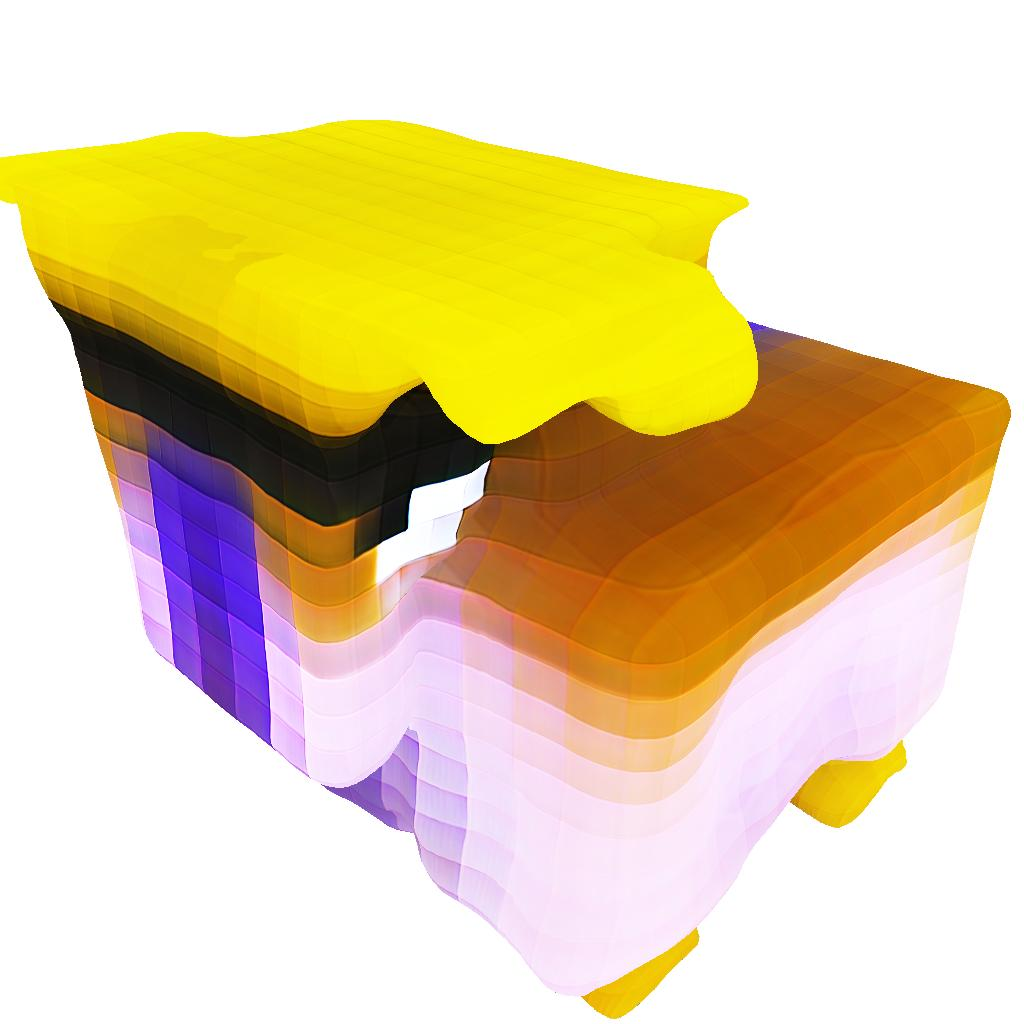} \\

\end{tabular}
    \vspace{-0.01\textheight}
    \caption{Qualitative comparison of the impact of %
    blending factor $\alpha$, where $\alpha = 0$ corresponds to no regularization.
}
    \label{fig:blend_compare_stylized}
\end{minipage}
    \vspace{-0.02\textheight}
\end{figure}

\vspace{-3mm}
\paragraph{Ablation on using stage-2 refinement.}
Stage-2 refinement gives large improvements in texture details over stage-1 training. We show the comparison in Fig.~\ref{fig:abaltion_stage}.

\vspace{-7mm}
\paragraph{Regularization}
We now investigate the design choice of how to weight $\blend$ between the SDS loss and the regularization loss as in Eq.~\ref{eq:blending}. %
In Tab.~\ref{tab:aba_method}, we show that adding regularization loss improves the quantitative results.
Here we qualitatively contrast training on different fixed weightings in Fig.~\ref{fig:blend_compare_stylized} and Appendix, respectively. 
We see greater consistency with the input shape by increasing shape regularization with higher blend factors $\blend$.

\begin{table} %
\vspace{-0.02\textheight}
\centering
    \caption{An ablation of unseen DreamFusion inputs in inference and annealing in training. Users prefer the annealed model with a dummy input in $51.2\%$ of cases.}
    \label{tab:point_cloud_input_inference} 
    \vspace{-0.01\textheight}
    \setlength{\tabcolsep}{2pt}    
    \begin{tabular}{l|ll|l}
    \toprule
    Model          & \multicolumn{2}{l}{without anneal} & {with anneal} \\
    Input               & Top-1    & Dummy         & Dummy          \\ 
    \hline
    CLIP     & 0.2628     & 0.2577          & 0.2605      \\
    FID           & 186.46 & 193.28 & 190.00 \\
    \bottomrule
    \end{tabular}
    
\vspace{-0.04\textheight}
\end{table}

\begin{figure}
    \vspace{-.3cm}
    \centering
    \includegraphics[width=0.95\linewidth]{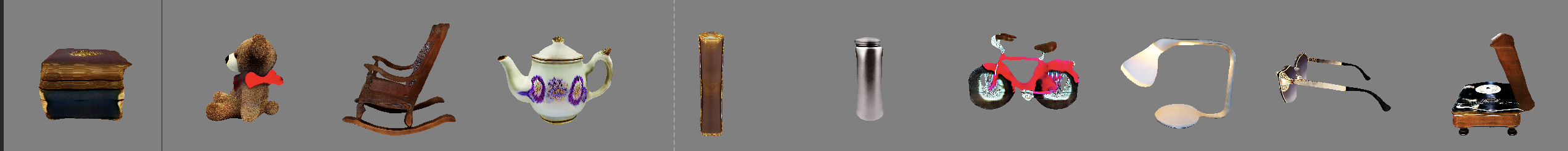}
    \includegraphics[width=0.95\linewidth]{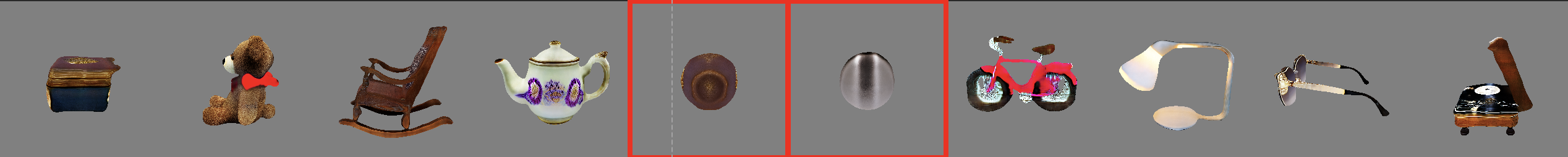}
    \includegraphics[width=0.95\linewidth]{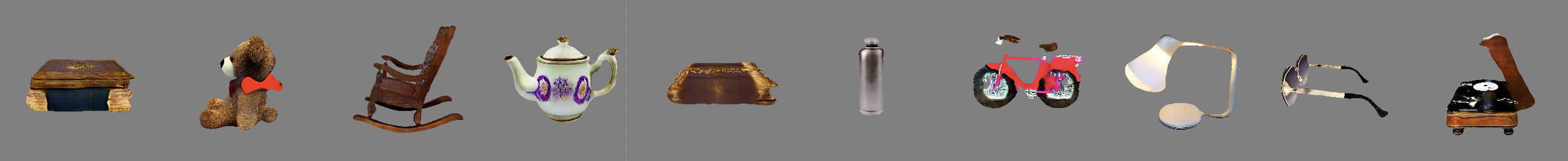}
    \vspace{-.1cm}
    \caption{
        Qualitative comparison of LATTE3D samples with a retrieved (top row) or dummy point cloud (middle row) before annealing. The performance is very similar except for a few geometry failures that inherit the dummy sphere geometry (highlighted in {\color{red}red}). We also show samples with dummy input after annealing (bottom row). After annealing, the model no longer exhibits this behavior on the dummy sphere inputs.
    }
    \label{fig:point_cloud_input}
    \vspace{-1.0cm}
\end{figure}

\paragraph{Annealing in training and different input point clouds in inference.} 
In training, we take retrieved point clouds from the dataset as input and anneal the input to become a fixed dummy point cloud. In inference, we only use the dummy point cloud as input. 
In Tab.~\ref{tab:point_cloud_input_inference}, we quantitatively ablate the role of this point cloud annealing process on the behavior of LATTE3D on \largeDataset. 
Our model without annealing training is somewhat sensitive to the point cloud input. There is a small performance drop when using a dummy as input in inference compared to a retrieved point cloud. 
However, the performance gap is reduced by introducing point cloud annealing so the model trains with dummy input. 
Further, we display qualitative results in Fig.~\ref{fig:point_cloud_input}, showing that point cloud annealing improves the results when the dummy input is used in inference. 

\vspace{-10mm}
\section{Conclusion and Limitation}
\vspace{-3mm}
We presented a scalable approach to perform amortized  text-to-enhanced-3D generation. %
To successfully scale amortization to larger datasets, we used 3D data through (1) 3D-aware 2D SDS, (2) pretraining, and (3) 3D regularization of the amortization process. We further improve the model architecture to be more scalable. Our model generates high-quality shapes within \generateTime. Moreover, quality and generalization can be improved further via a speedy test time optimization. 
Our model also has limitations. First, our model uses SDS and thus relies on the understanding of the text-to-image models, which can often fail to respect fine-grained details such as part descriptions. Second, the geometry is frozen in stage-2 and test-time optimization, so the geometry flaws from stage-1 cannot be fixed. Developing amortized training for stable geometry changes in stage-2 is left for future work.

\newpage

\section*{Acknowledgements}\label{sec:ack}
    \noindent
    We thank Matan Atzmon, Or Perel, Clement Fuji Tsang, Masha Shugrina, and her group for helpful feedback.
    The Python community ~\cite{van1995python, oliphant2007python} made underlying tools, including PyTorch~\cite{paszke2017automatic} \& Matplotlib~\cite{hunter2007matplotlib}.

\section*{Disclosure of Funding}
    \noindent
    NVIDIA funded this work.
    Kevin Xie, Jonathan Lorraine, Tianshi Cao, Jun Gao, and Xiaohui Zeng had funding from student scholarships at the University of Toronto and the Vector Institute, which are not in direct support of this work.

{
    \bibliographystyle{splncs04}
    \bibliography{main}
}

\clearpage
\appendix

\title{\vspace{0.5em}Supplementary Material\\\vspace{1.0em}\titleString}
\titlerunning{\titlerunningString}
\author{\authorString}
\authorrunning{\authorrunningString}
\institute{\instituteString}
\maketitle

\noindent
In this appendix, we provide additional details on \ours.
In Sec.~\ref{sec:app_methodology}, we provide details on our model architecture, amortized learning framework, and test time optimization.
In Sec.~\ref{sec:app_experiments}, we provide details on the experimental settings, additional results of our methods, comparisons with baselines, and additional ablation studies.
In Sec.~\ref{sec:app_failure}, we show the failure cases of our method and its limitations.
In Sec.~\ref{sec:app_amort_reg}, we show an additional capability for amortizing the regularization strength, allowing enhanced user-controllability and interpolations. 

\renewcommand{\thetable}{A.\arabic{table}}
\begin{table}[h]\caption{Glossary and notation}
    \vspace{-0.04\textheight}
    \begin{center}
        \begin{tabular}{c c}
            \toprule
            ATT3D & Amortized Text-to-3D~\cite{lorraine2023att3d}\\
            SDS & Score Distillation Sampling~\cite{poole2022dreamfusion}\\
            SD & Stable Diffusion\\
            SDF & Signed distance field\\
            MV & Shorthand for MVDream~\cite{shi2023MVDream}\\
            PT & Pretraining\\
            S1, S2 & Stage-1 and stage-2\\
            FID & Fr\'echet Inception Distance\\
            OOD & Out of distribution\\
            MLP & Multilayer Perceptron\\
            $\inputText$ & A text prompt\\
            $\inputShape$ & An input shape point cloud\\
            $\geometryNet$ & The geometry network in our model \\
            $\textureNet$ & The texture network in our model \\
            $\modelF=(\geometryNet,\textureNet)$ & Our model to generate 3D objects\\
            $\outputShape = \modelF(\inputText, \inputShape)$ & A predicted output shape\\
            
            $\camera$ & A sampled camera\\
            $\renderer(\outputShape, \camera)$ & A rendering function\\
            $\renderer_{\text{opacity}}, \renderer_{\text{RGB}}$ & A renderer for the opacity or RGB\\
            $\lossReg(\outputShape, \inputShape, \camera)$ & The shape regularization loss using only opacity\\
            $\lossInit(\outputShape, \inputShape, \camera)$
            & The pretraining reconstruction loss using RGB\\
            $\lossText$ & The SDS loss for our text-prompt\\
            $\blend \in [0, 1]$ & Blend factor between regularization and SDS\\
            $\lossTrain = (1 - \blend) \lossText + \blend \lossReg$ & The training loss \\
            $\modelF'(\inputText, \inputShape, \blend)$ & Our model for amortizing over blend factors \\
            \bottomrule
        \end{tabular}
    \end{center}
    \label{tab:TableOfNotation}
    \vspace{-0.03\textheight}
\end{table}
\renewcommand{\thefigure}{\thesection.\arabic{figure}}
\renewcommand{\thetable}{\thesection.\arabic{table}}

\section{Additional Methodology Details}
\label{sec:app_methodology}
In this section, we provide additional details on the model architecture for both geometry and texture network (Sec.~\ref{sec:app_architecture}), with more details on the amortized learning in Sec.~\ref{sec:app_details_amortized} and test time optimization in Sec.~\ref{sec:app_test_optim}.

\subsection{Background on Amortized Learning}
\label{sec:app_background_amort}
Amortized optimization methods predict solutions when we repeatedly solve similar instances of the same problem~\cite{amos2022tutorial}, finding use in generative modelling~\cite{kingma2013auto, rezende2014stochastic, cremer2018inference, wu2020meta}, hyperparameter optimization~\cite{lorraine2018stochastic, mackay2019self} and more.
The goal is to solve the problem more efficiently by sharing optimization efforts somehow.

A typical amortized optimization strategy involves (a) finding some context or description of the problem you are solving, (b) making that an input to a network, and (c) training the network to output problem solutions given an input problem context, by sampling different problems during training.
Specific methods differ in how they construct the problem context, the architecture for inputting the context to the network, and how they sample the context during training.
This strategy circumvents the need to completely solve any single problem, scaling more efficiently than na\"ively generating a large dataset of solved problems and training a network to regress them.
Many feedforward networks can be viewed as amortizing over some type of problems, but the amortized viewpoint is particularly useful when the target associated with the network input is the result of (an expensive) optimization.

We amortize the optimization of text-to-3D methods, which repeatedly solve an expensive optimization problem to generate a 3D object consistent with a provided text prompt using the SDS loss.
The problem context is the text prompt embedding, and the solution is the output 3D object.

\subsection{Model Architecture}
\label{sec:app_architecture}

\begin{figure*}[h!]
    \includegraphics[width=\linewidth,trim={0 1cm 10cm 0cm},clip]{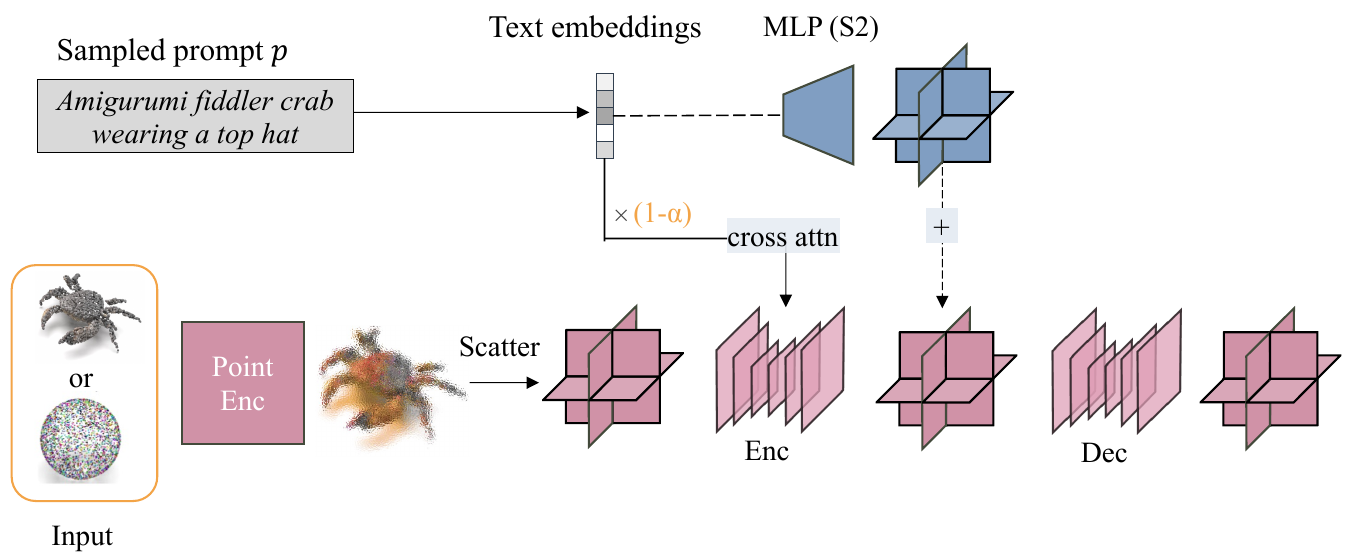}
    \caption{We show a pipeline diagram with our architecture's individual components.
    Note that the additional residual MLP is only active in the texture network upsampling in stage-2.
    Dummy point cloud }
    \vspace{-0.01\textheight}
    \label{fig:app_pipeline}
\end{figure*}

We include an architecture diagram showing the individual components in our model in Fig.~\ref{fig:app_pipeline}.

\subsubsection{Geometry Network}
\label{sec:app_geo_architecture}
We compute the token embeddings from the input text prompt using a frozen CLIP~\cite{clip} text encoder, following Stable Diffusion 2.1~\cite{Rombach_2022_CVPR}.
The tokens are padded or truncated to $77$ for easy batching.
In addition, we uniformly sample $\num{20480}$ points from the reference input 3D shape and feed in the point's positions and RGB values to the network.

We adopt PointNet~\cite{qi2017pointnet} as our point cloud encoder but use triplanes for local pooling. 
We use scatter operations at every block to aggregate the point features onto the triplane via geometric projection. This can be regarded as a local pooling among the points projected into the same pixel in the triplane. We then gather the feature from the triplane and concatenate it with the initial point feature for each point as input to the residual stream MLP.

We use $128\times128$ resolution triplanes for our pooling and the same resolution for the output triplane feature in the point encoder.
In our experiments, the point encoder has $5$ residual blocks and a hidden dimension of $32$.
The output of our point encoder is a $128\times128$ triplane with $32$ channels obtained after a final projection of the point features.

After obtaining the point encoder's triplane feature, we use a U-Net (encoder) to combine the features with the text embeddings, followed by another U-Net (decoder) to process the triplane features.
The encoder and decoder U-Nets are nearly identical, except the encoder conditions on the text embeddings via cross-attention layers inserted after every U-Net residual block.
Specifically, in our residual blocks, we apply a $3\times3$ rolled convolution (convolution is applied to height-stacked triplanes) followed by $1\times1$ 3D-aware convolution~\cite{Wang2022RODIN,gupta20233dgen} and another $3\times3$ rolled convolution.
We use Swish activation and GroupNorm with $32$ groups throughout.
Our U-Net has a depth of $5$, and the lowest resolution is $8$.
We use a single res block per depth, and the base number of channels is $64$.
The final output of the U-Net is a $128$ resolution, $32$ channel triplane feature.

In the encoder, the cross-attention layer uses the flattened triplane features as the query and the text token embeddings as context. It uses multi-head attention with $4$ heads, a head dimension of $64$, and layernorm.

We first compute a feature by summing bilinearly interpolated features from each feature plane in the triplane to predict the SDF for each query position. We then pass this feature vector through an MLP network to obtain an approximate SDF value. In our implementation, we add the predicted SDF value with the SDF value for a sphere, stabilizing the training process.
The final SDF value will be converted to density for volume rendering or isosurface extraction for surface-based rendering.
The MLP network has one hidden layer of size $32$.

\subsubsection{Texture Network}
The texture network $\textureNet$ has the same architecture as the geometry network described in Sec.~\ref{sec:app_geo_architecture}.
However, we have an upsampling layer to increase the triplane resolution to learn high-frequency texture details in stage-2. We describe the details for this upsampling layer below.

\paragraph{Texture network upsampling}
We first upsample the triplane from a resolution of $128$ to $480$ using nearest-neighbors and add a residual MLP that maps the CLIP embedding to the latent triplane, following the \emph{mapping network} from  ATT3D~\cite{lorraine2023att3d}. 
Specifically, with the input prompt text-embedding $\inputText$ and output triplane features $\latentTriplane$ we use:
\begin{equation}
    \mappingNetEmbedding = \smash{\textnormal{linear}_{\textnormal{w/ bias}}}\left(\textnormal{normalize}\left(\tanh{\left(0.1 \cdot \textnormal{flatten}\left(\inputText\right)\right)}\right)\right)
\end{equation}
\begin{equation}
    \latentTriplane = \textnormal{reshape}\left(\smash{\textnormal{linear}_{\textnormal{no bias}}}\left(\mappingNetEmbedding\right)\right)
\end{equation}
Here, $\mappingNetEmbedding$ is of size $32$, and the final layer uses no bias, as in ATT3D.  We found spectral normalization was not required, and thus, we discarded it in our model. The $\textnormal{reshape}$ operation takes a flattened vector to our $32$ channel feature triplane shape of $[3, 32, 480, 480]$. We use an activation of $\tanh(0.1 \cdot)$ on the text-prompt embedding to scale the entries to $(-1, 1)$ for optimization stability, where $0.1$ is to prevent saturation. We also normalized each prompt embedding's activation components to have mean $0$ and unit variance for optimization stability.

The texture network is then refined via the second-stage optimization using surface renderings.
We find that upsampling the geometry triplane leads to unstable training,  we only upsample the texture network and freeze the geometry in the second stage.

\subsection{Details on Amortized Learning}
\label{sec:app_details_amortized}

\paragraph{Pretraining.}
For pretraining with reconstruction, we use surface-based rendering for its efficiency. We train the model using a batch size of $64$ objects and render $2$ random views per object, which gives us $128$ total views.

\paragraph{Stage-1 training.}
In stage-1, we use $4$ training views per output shape, as MVDream~\cite{shi2023MVDream} only supports computing SDS loss on $4$ views. We use $8$ shapes per batch totalling $32$ views for all experiments except for \largeDataset where we increase the batch size by a factor of $4$ to $32$ shapes. 
We use $96$ samples per ray for volume rendering to render an image at a resolution of $\num{256}\times\num{256}$.
We also render one extra view per shape to compute our shape regularization loss.

\paragraph{Input point cloud annealing} As in the main text, towards the end of stage-1, we anneal away the dependence on the input point clouds by gradually replacing them with the dummy point cloud. Specifically, we randomly replace a subset of points in the input point cloud with points from the dummy point cloud using a probability that increases linearly from $0$ to $1$ in the last $5000$ training iterations. During this phase, we also freeze the decoder network to only modify the point cloud encoder and encoder U-Net. This puts additional constraints on the output shapes to not change much in appearance from before annealing.

\paragraph{Stage-2 training.}
In stage-2, we randomly sample $10$ training views per shape, with $8$ output shapes per batch, totaling $80$ views for all experiments. 
To render the image, we first use differentiable marching cubes with a grid resolution size of $256$ and render images with a resolution $\num{1024}\times\num{1024}$ which is bilinearly downsampled to $512$ for our image-denoising steps.

\paragraph{Hyperparameters.}
We optimize $\textureNet$ for $600$ iterations to reduce runtime and anneal the timestep sampling range for the SDS loss over the first $400$ iterations. Further iterations do not bring noticeable improvements in our experiments.
We use the Adam optimizer with a $\num{0.0003}$ learning rate and $\beta_1=\num{0.9}$, $\beta_2=\num{0.99}$. 
We apply a $\num{0.001}$ multiplier when computing the text guidance loss ($\lossText$) to scale it with our shape regularization loss ($\lossReg$). Different amounts of shape regularization may be optimal depending on the alignment between the input reference shapes and the associated prompts.
For our main reported results, we used $\alpha=0.2$ for \animalReal, $\alpha=\num{0.1}$ for \animalStyle and $\alpha=\num{0.05}$ for \largeDataset.

\subsection{Test Time Optimization}
\label{sec:app_test_optim}
For test-time optimization, we decrease the learning rate to $\num{0.0001}$.
We anneal the sampled timestep range from $[0.5,0.9]$ to $[0.1,0.5]$ over $400$ iterations.
We sample $8$ training views per iteration and one $\epsilon$ noise sample per training view.
Other settings, such as architecture and guidance, are the same as in stage-2. 

\section{Additional Experiments}
\label{sec:app_experiments}
In this section, we provide additional experiments. We first provide the details on our experimental setup in Sec.~\ref{sec:app_exp_setup}.

\subsection{Experimental Setup Details}
\label{sec:app_exp_setup}
\subsubsection{Details on Baselines}

\noindent
\paragraph{\textbf{MVDream.}} We use an open-source implementation of MVDream~\cite{shi2023MVDream} from threestudio~\cite{threestudio2023}, with the original (publicly available) $4$-view diffusion model. We follow the default settings from threestudio, which we find to perform well. We fit each 3D model using SDS for $\num{10000}$ training iterations, with a batch size of $8$ on a single GPU.

\noindent
\textbf{ATT3D.}
We re-implement ATT3D~\cite{lorraine2023att3d} as a baseline in our experiments.
We use their amortized optimization with either (a) the original hypernetwork or (b) our triplane architecture.
Our reproduction is consistently better than the original paper, with more detailed results in Sec.~\ref{sec:app_validate_att3d}.
Our architecture includes a triplane latent input representation instead of a multi-resolution voxel grid, as in ATT3D.
We also have a point-cloud encoder in our architecture -- absent in ATT3D -- that feeds a dummy, spherical point cloud for a fair comparison.

\noindent
\textbf{3DTopia} We use the first stage of the open-source implementation from \url{https://github.com/3DTopia/3DTopia}. We use the default settings for sampling the diffusion model (DDIM sampler with $200$ steps, guidance scale $7.5$. We decode and volumetrically render the sampled triplane into $256$ by $256$ images for the user study. Note that the $20$ seconds compute time only includes the compute cost of the diffusion sampling, as time for rendering the triplane depends both on the number of camera poses and the desired resolution.

\noindent
\textbf{LGM} We use the open-source implementation from \url{https://github.com/3DTopia/LGM}. We use the text-to-3D pipeline implemented in the code base. To sample multi-view images from MVDream, we use the default settings in the LGM code base (DDIM sampler with $30$ steps, guidance scale $7.5$). Using the LGM repository, we render the generated Gaussian Splatting for the user study. As with 3DTopia, the $6$ seconds compute time only includes sampling MVDream and running the Gaussian reconstruction model, though the per-frame LGM rendering time is generally negligible compared to 3DTopia.

\subsubsection{Dataset}

\begin{table*}
\small
    \centering
    \caption{We show details for various amortized datasets, where the source generated the prompts.
    Each row is one curated dataset, with our assigned name for each dataset.
    Each prompt is tied to one or multiple shapes.
    The last column shows example prompts that are tied to the same training shape.}
    \begin{tabularx}{\textwidth}{c c c c >{\raggedright\arraybackslash}p{5.5cm}} %
       \toprule
       \textbf{name} & \textbf{\#shape} & \textbf{\#prompts} & \textbf{source} & \textbf{prompt examples} \\ 
       \midrule
     \animalReal & 100  & 1,000 & ChatGPT & (1) \textit{Gray wolf} (2) \textit{Arctic wolf} (3) \textit{Coyote} (4) \textit{Dingo}  (5) \textit{German shepherd dog} (6) \textit{Siberian husky} (7) \textit{Alaskan malamute} (8) \textit{Fox} %
       \\ \addlinespace
       \animalStyle & 100  & 12,000 & Rules & (1) \textit{Penguin wearing a top hat animal built out of lego} (3) \textit{wood carving of penguin in a Chef’s Hat} \\ \addlinespace
       \largeDataset & 33,958  & 101,608 & ChatGPT & 
       (1) \textit{A sleek silver dumbbell with a digital display that shows the weight and number of reps} (2) \textit{A futuristic dumbbell made of transparent glass with swirling neon lights inside} 
       (3) \textit{A chrome-plated dumbbell with a textured grip surface for better hold} 
       \\ 
       \bottomrule
    \end{tabularx}
    \label{tab:training_data_appendix}
\end{table*}
\paragraph{Curating prompt sets.}
To create our prompt sets, we start with captions for the objects in our shape set~\cite{luo2023scalable}.
We then use GPT or a rules-based approach to create multiple prompts based on the caption.

Specifically for the \animalReal dataset, we ask GPT to recommend $10$ other animals with the same coarse shape.
We found it important to add additional instructions to ensure diverse responses.
The instruction used was:
\textit{``Based on the animal named in the query, provide $10$ different visually distinct animal species or breeds that have roughly a similar body shape (they can be different scale). Start from similar animals and gradually make them much less similar. The last $4$ should be very different. Use common name and do not include esoteric variants. Keep your answers short and concise."}

For constructing \largeDataset, we broaden the scope of the instruction to handle general object types and gather $3$ prompts per caption. The instructions used were:
\textit{``I am using text conditioned generative models, it take a base shape + target prompt as input, and deforms the shape to a similar object, with a slightly different shape / appearance. Now I have a caption describing the input object, imagine the shape of the object, and think about $3$ possible target prompts that this object can deform into? Make the prompt as detailed as possible. Attention: 1) Don't describe the background, surroundings. 2) output is a python list"}

\paragraph{Curating shape datasets.}
We found undesirable shapes in the Objaverse~\cite{objaverse} dataset, such as those resulting from scans or abstract geometry, which we remove in a semi-manual process.  Specifically, we first cluster shapes into $\num{1000}$ clusters through K-means using the average multiview CLIP image embeddings. We then manually filter the clusters that have undesirable shapes. This gives us 380k shapes in total. To get the reference shapes for $\largeDataset$, we take the intersection between our filtered subset and the subset with lvis categories released by Objaverse~\cite{objaverse}. For shapes in $\animalReal$ and $\animalStyle$ datasets, we manually pick $100$ animals with an acceptable coarse full-body geometry and represent a diversity of animals. %

\paragraph{Data pre-processing.}
We use Blender~\cite{blender} to preprocess the dataset from Objaverse. We first normalize all shapes to be centered, with the same maximum length of $\num{0.8}$. We then randomly select $32$ cameras on an upper sphere and render $32$ images for each shape. To obtain the point cloud, we render the depth images and unwarp them into 3D to obtain the point cloud with colors.

\subsubsection{Evaluation Metrics}
\paragraph{Render-FID.}
The Render-FID is computed as the FID score between the images generated by Stable Diffusion 2.1 and those rendered from our model's generated shapes.  Specifically,
for each prompt in the test set, we first augment them with $4$ view-dependent prompts, and then for each augmented prompt, we use Stable Diffusion 2.1 to generate $4$ different images. 
For our rendered images, we generate one shape from one text prompt and render four images for each generated shape from $0$, $90$, $180$, and $270$ degrees. 

\paragraph{CLIP scores.}
We compute the average CLIP scores between the text prompt and each rendered image to indicate how well the generated 3D shape aligns with the input text prompt. 
It is computed between the CLIP embeddings of the input text prompt and renderings from our output shape and averaged over multiple views.
We compute the CLIP score by using the \newline\verb|laion/CLIP-ViT-H-14-laion2B-s32B-b79K| version of the CLIP model.

\vspace{-0.01\textheight}
\paragraph{Mask-FID}
We additionally introduced the Mask-FID metric to quantitatively assess the adherence of our geometry to the shape dataset. We compute the FID between binary masks of the generated shapes and those of the 3D dataset.
In particular, we take $100$ input shapes and render binary masks from them from $4$ $90$ degree rendering views.
We compare these with binary masks rendered from $4$ views of our output shapes.
We threshold the opacity at $0.5$ for volume rendering to obtain the mask.

\vspace{-0.01\textheight}
\paragraph{User study.} 
We use Amazon Mechanical Turk to operate a user study. We present two videos from two different methods for each user with the corresponding text prompt. One of the videos is from the baseline, and the other is from our method. We ask the user to answer six questions:
\begin{itemize}
    \item Which 3D object has a more natural visual appearance?
    \item Which 3D object has a visual appearance with more details?
    \item Which 3D object has a more natural shape?
    \item Which 3D object has a shape with more geometric details?
    \item Which 3D object has an overall more appealing look?
    \item Which 3D object more faithfully follows the textural description?
\end{itemize}

\begin{figure*}%
\includegraphics[width=\linewidth,trim={0 0 0 0},clip]{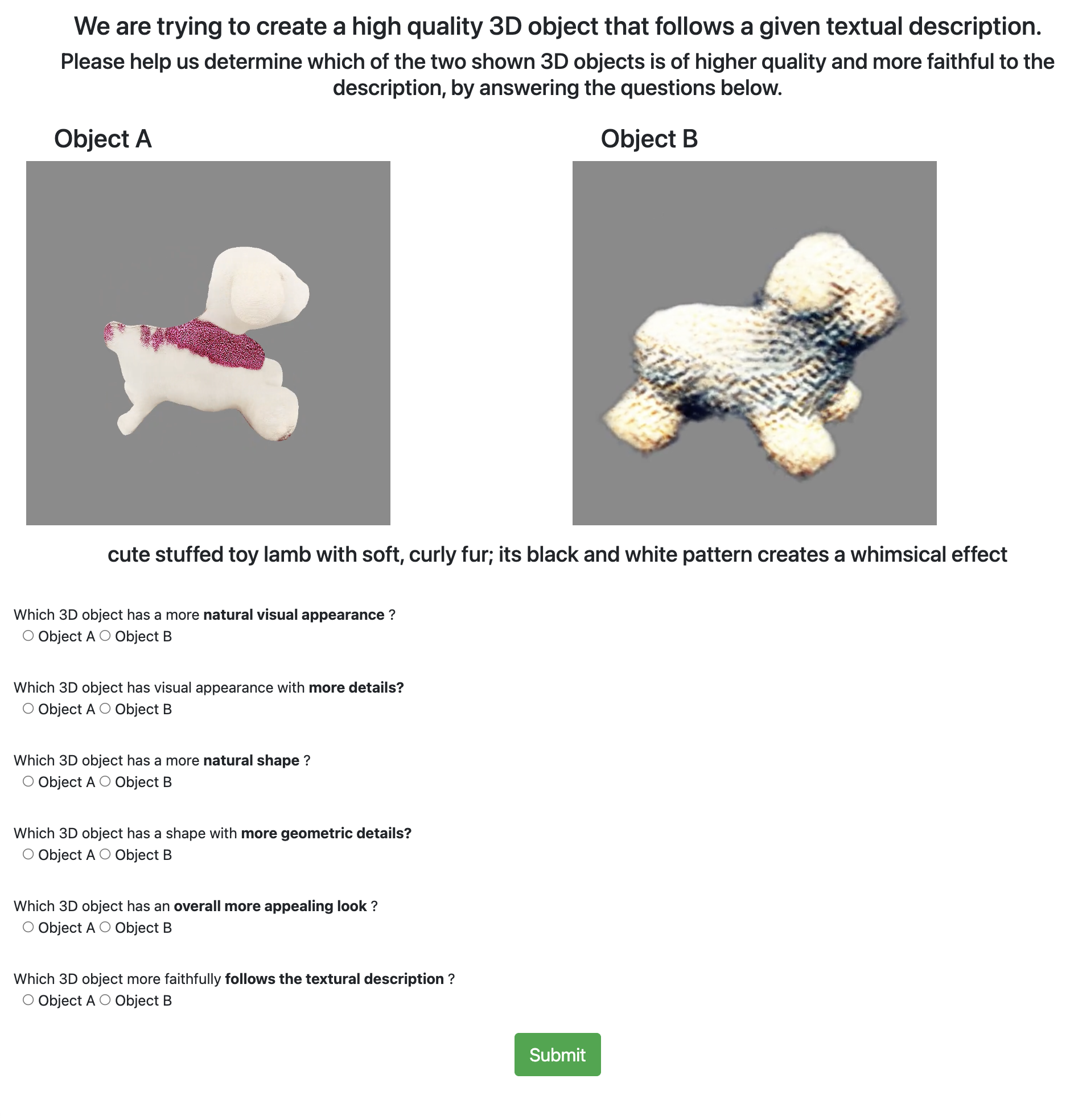}
    \caption{
        We show a screenshot of one user study we conducted in our experiment. 
        The visualizations for objects A and B were displayed as a video of a camera sweeping over the azimuth.
    }
    \label{fig:sup_user_study_img}
\end{figure*}

We provide a screenshot of the user study in Fig.~\ref{fig:sup_user_study_img}. We use $200$ prompts per method, and each comparison is shown to three different users to reduce variance. We randomize the order of the videos to remove bias due to ordering.

\subsection{Validating ATT3D Reproduction}\label{sec:app_validate_att3d}
We assess our reproduction of the ATT3D baseline method with qualitative comparisons by training models on DF27, the $27$ prompts from DreamFusion re-used in ATT3D.
Fig.~\ref{fig:df27reimpl} shows that our baseline with a triplane+UNet architecture and only stage-1 training achieves a quality similar to that of the source ATT3D.

Furthermore, we see improvements with stage-2 amortized refinement (without 3D data), dubbed \emph{MagicATT3D} or ATT3D+S2, as we introduce amortizing over the refinement stage of Magic3D~\cite{lin2023magic3d}.
Fig.~\ref{fig:magicatt3d} shows the quality improvement from stage-1 to stage-2 is comparable to Magic3D's quality improvement over single-stage methods like DreamFusion. These experiments demonstrate the successful reproduction of ATT3D in our framework. 

\input{images/reimplement_df27}

\subsection{Additional Results}

We provide additional results for the three datasets we utilized in our main paper. For \animalStyle and \largeDataset, we provide additional qualitative examples of our model output for unseen prompts, highlighting our generalization ability.

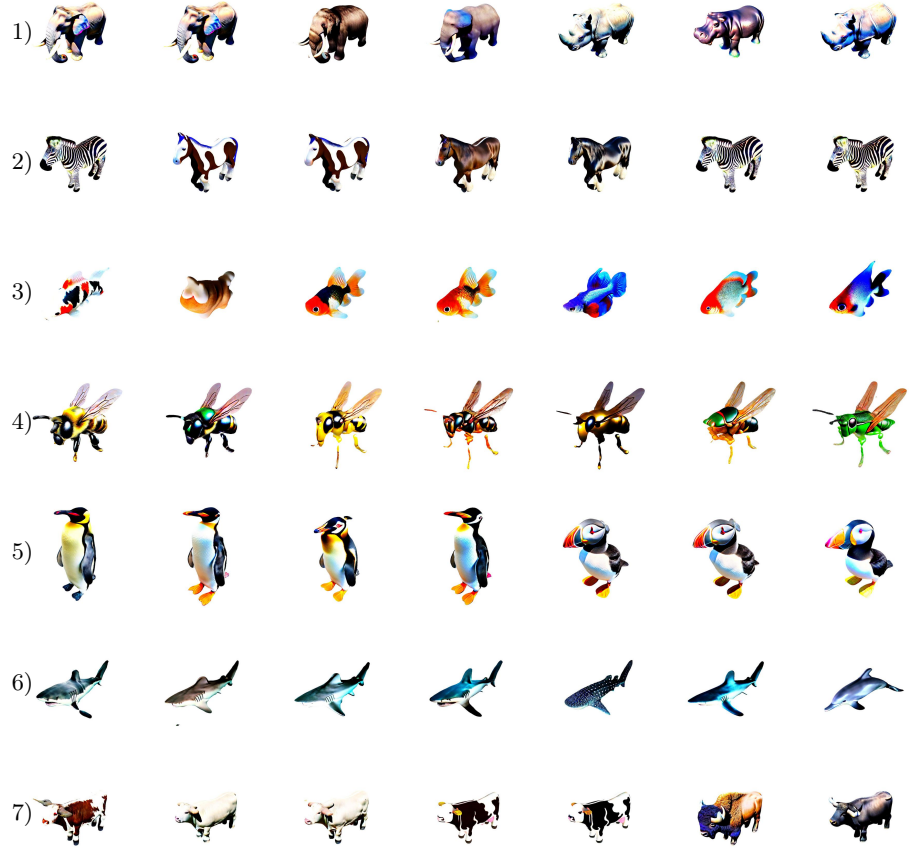
\begin{figure*}[p]%
    \begin{tikzpicture}
        \node (img11){\includegraphics[width=\linewidth,trim={0 0 0 0},clip]{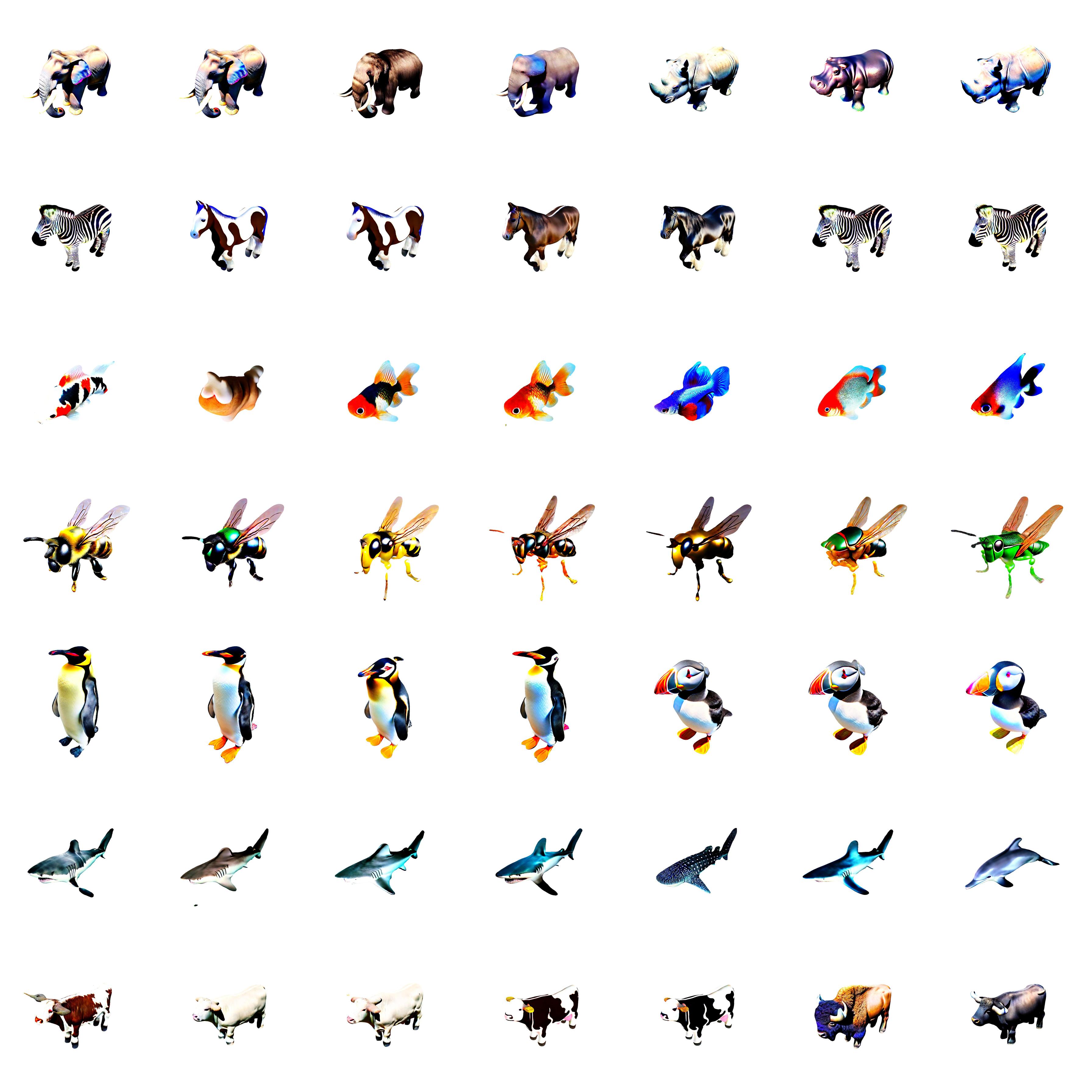}};
        \node[left=of img11, node distance=0cm, rotate=0, xshift=1.5cm, yshift=5.2cm,  font=\color{black}]{1)};
        \node[left=of img11, node distance=0cm, rotate=0, xshift=1.5cm, yshift=3.47cm,  font=\color{black}]{2)};
        \node[left=of img11, node distance=0cm, rotate=0, xshift=1.5cm, yshift=1.73cm,  font=\color{black}]{3)};
        \node[left=of img11, node distance=0cm, rotate=0, xshift=1.5cm, yshift=0.0cm,  font=\color{black}]{4)};
        \node[left=of img11, node distance=0cm, rotate=0, xshift=1.5cm, yshift=-1.73cm,  font=\color{black}]{5)};
        \node[left=of img11, node distance=0cm, rotate=0, xshift=1.5cm, yshift=-3.47cm,  font=\color{black}]{6)};
        \node[left=of img11, node distance=0cm, rotate=0, xshift=1.5cm, yshift=-5.2cm,  font=\color{black}]{7)};

    \end{tikzpicture}
    \caption{
        We show additional example results of \ours on the seen prompts from \animalReal. The rows represent generations based on the same point cloud input. For each row the animal in the text prompts are:
        \textit{
        1) Asian Elephant; Sumatran Elephant; Woolly Mammoth; Mastodon; White Rhinoceros; Hippopotamus; Indian Rhinoceros
        2) Zebra; Paint Horse; Pinto Horse; Brindle Horse; Friesian Horse; Grevy's Zebra; Mountain Zebra
        3) Koi ; Shubunkin; Oranda Goldfish; Fantail Goldfish; Betta Fish; Gourami; Tetra Fish
        4) Honey bee; Carpenter bee; Yellow jacket; Paper wasp; Hornet; Cicada ; Grasshopper
        5) Emperor Penguin; Gentoo Penguin; Macaroni Penguin; Rockhopper Penguin; Puffin; Atlantic Puffin; Tufted Puffin
        6) Great White Shark; Tiger Shark; Bull Shark; Mako Shark; Whale Shark; Blue Shark; Dolphin
        7) Texas Longhorn; Charolais Bull; Brahman Bull; Hereford Bull; Simmental Bull; Bison; Water Buffalo}
        }
    \label{fig:qualitative_grid_real}
\end{figure*}

\vspace{-0.015\textheight}
\paragraph{Additional results on \animalReal.}
We present additional qualitative examples in Fig.~\ref{fig:qualitative_grid_real}.

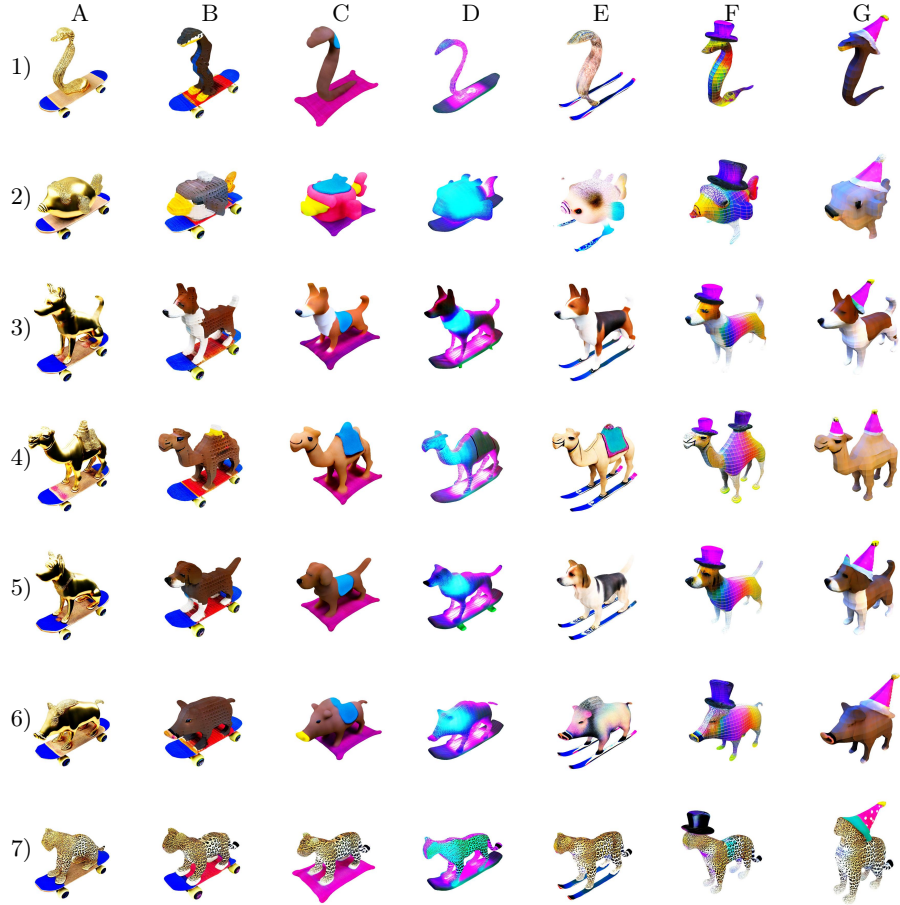
\begin{figure*}%
    
    \begin{tikzpicture}
        \node (img11){
        \includegraphics[width=\linewidth,trim={0 0 0 0},clip]{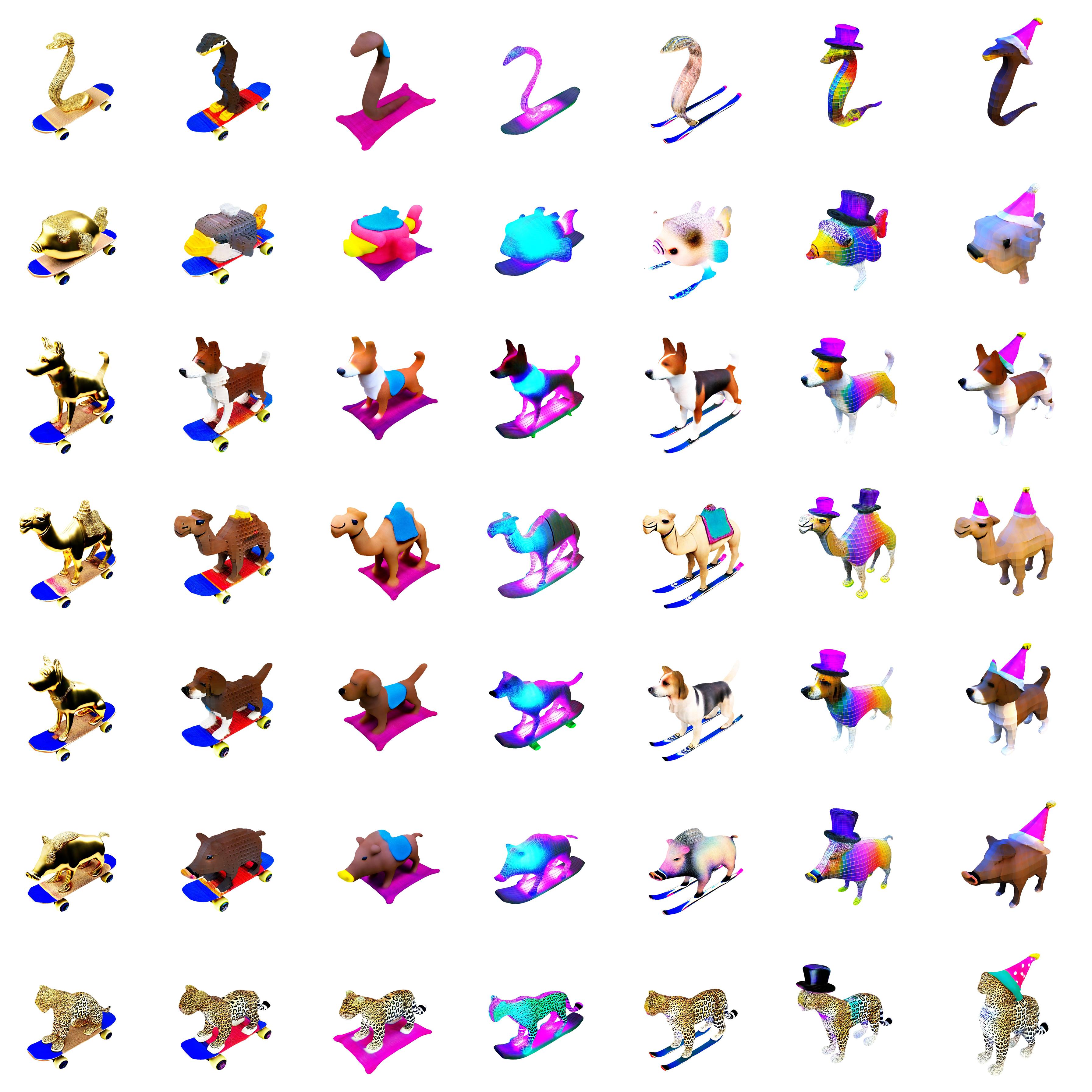}};
        \node[left=of img11, node distance=0cm, rotate=0, xshift=1.5cm, yshift=5.2cm,  font=\color{black}]{1)};
        \node[left=of img11, node distance=0cm, rotate=0, xshift=1.5cm, yshift=3.47cm,  font=\color{black}]{2)};
        \node[left=of img11, node distance=0cm, rotate=0, xshift=1.5cm, yshift=1.73cm,  font=\color{black}]{3)};
        \node[left=of img11, node distance=0cm, rotate=0, xshift=1.5cm, yshift=0.0cm,  font=\color{black}]{4)};
        \node[left=of img11, node distance=0cm, rotate=0, xshift=1.5cm, yshift=-1.73cm,  font=\color{black}]{5)};
        \node[left=of img11, node distance=0cm, rotate=0, xshift=1.5cm, yshift=-3.47cm,  font=\color{black}]{6)};
        \node[left=of img11, node distance=0cm, rotate=0, xshift=1.5cm, yshift=-5.2cm,  font=\color{black}]{7)};
        \node[above=of img11, node distance=0cm, rotate=0, xshift=-5.2cm, yshift=-1.5cm,  font=\color{black}]{A};
        \node[above=of img11, node distance=0cm, rotate=0, xshift=-3.47cm, yshift=-1.5cm,  font=\color{black}]{B};
        \node[above=of img11, node distance=0cm, rotate=0, xshift=-1.73cm, yshift=-1.5cm,  font=\color{black}]{C};
        \node[above=of img11, node distance=0cm, rotate=0, xshift=0cm, yshift=-1.5cm,  font=\color{black}]{D};
        \node[above=of img11, node distance=0cm, rotate=0, xshift=1.73cm, yshift=-1.5cm,  font=\color{black}]{E};
        \node[above=of img11, node distance=0cm, rotate=0, xshift=3.47cm, yshift=-1.5cm,  font=\color{black}]{F};
        \node[above=of img11, node distance=0cm, rotate=0, xshift=5.2cm, yshift=-1.5cm,  font=\color{black}]{G};
    \end{tikzpicture}
    \caption{
    We show additional example results of \ours on the seen prompts from \animalStyle. Each column shares the same style combination and each row is a different animal. From left to right, the style combinations are: %
    From left to right, the style combinations are:
    \textit{
    A) engraved golden cast sculpture of .. on top of a skateboard
    B) fun cute .. on top of a skateboard animal built out of lego pieces
    C) fun kids play doh .. on a Magic Carpet
    D) cyberpunk neon light .. on top of a snowboard 3D model
    E) highly detailed .. wearing skis painted in pastel delicate watercolors
    F) colorful 3D mosaic art suplture of .. wearing a top hat, unreal engine, dimensional surfaces
    G) .. wearing a party hat, low poly 3D model, unreal engine
    }
    From top to bottom, the animals are:
    1) cobra
    2) pufferfish
    3) basenji dog
    4) camel
    5) dog
    6) boar
    7) leopard
    }
    \label{fig:qualitative_grid_style_seen}
\end{figure*}

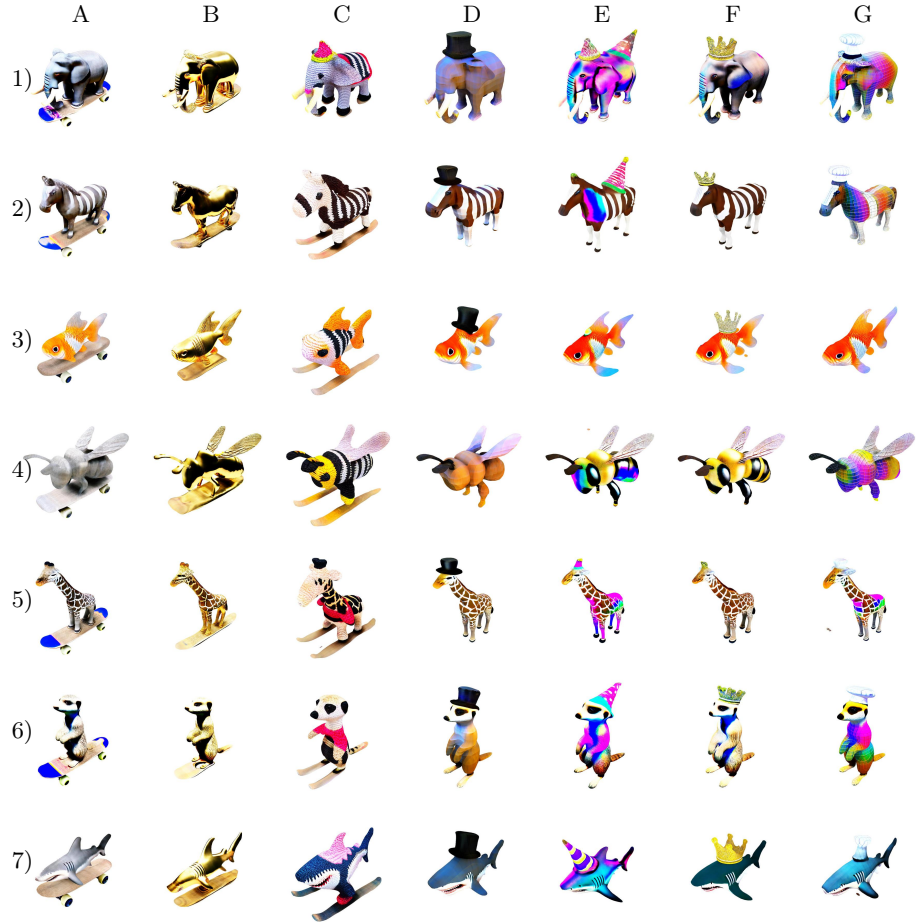
\begin{figure*}%
    
    \begin{tikzpicture}
        \node (img11){
        \includegraphics[width=\linewidth,trim={0 0 0 0},clip]{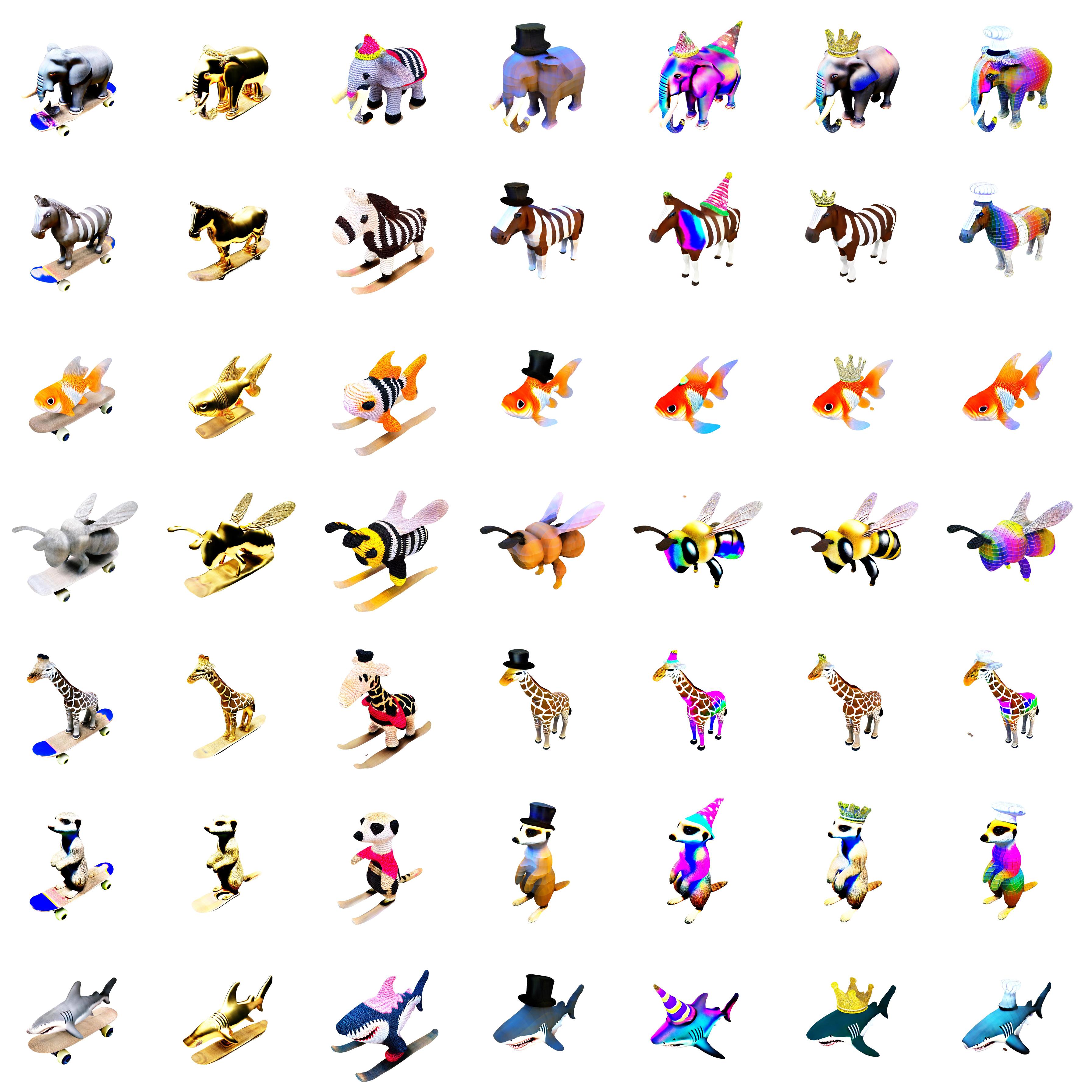}
        };
        \node[left=of img11, node distance=0cm, rotate=0, xshift=1.5cm, yshift=5.2cm,  font=\color{black}]{1)};
        \node[left=of img11, node distance=0cm, rotate=0, xshift=1.5cm, yshift=3.47cm,  font=\color{black}]{2)};
        \node[left=of img11, node distance=0cm, rotate=0, xshift=1.5cm, yshift=1.73cm,  font=\color{black}]{3)};
        \node[left=of img11, node distance=0cm, rotate=0, xshift=1.5cm, yshift=0.0cm,  font=\color{black}]{4)};
        \node[left=of img11, node distance=0cm, rotate=0, xshift=1.5cm, yshift=-1.73cm,  font=\color{black}]{5)};
        \node[left=of img11, node distance=0cm, rotate=0, xshift=1.5cm, yshift=-3.47cm,  font=\color{black}]{6)};
        \node[left=of img11, node distance=0cm, rotate=0, xshift=1.5cm, yshift=-5.2cm,  font=\color{black}]{7)};

        \node[above=of img11, node distance=0cm, rotate=0, xshift=-5.2cm, yshift=-1.35cm,  font=\color{black}]{A};
        \node[above=of img11, node distance=0cm, rotate=0, xshift=-3.47cm, yshift=-1.35cm,  font=\color{black}]{B};
        \node[above=of img11, node distance=0cm, rotate=0, xshift=-1.73cm, yshift=-1.35cm,  font=\color{black}]{C};
        \node[above=of img11, node distance=0cm, rotate=0, xshift=0cm, yshift=-1.35cm,  font=\color{black}]{D};
        \node[above=of img11, node distance=0cm, rotate=0, xshift=1.73cm, yshift=-1.35cm,  font=\color{black}]{E};
        \node[above=of img11, node distance=0cm, rotate=0, xshift=3.47cm, yshift=-1.35cm,  font=\color{black}]{F};
        \node[above=of img11, node distance=0cm, rotate=0, xshift=5.2cm, yshift=-1.35cm,  font=\color{black}]{G};

    \end{tikzpicture}
    \caption{
        We show additional example results of \ours on the unseen prompts from \animalStyle demonstrating combinatorial generalization. Each column shares the same style combination and each row is a different animal. 
        From left to right, the style combinations are:         
        \textit{
        A) gray textured marble statue of .. on top of a skateboard
        B) engraved golden cast sculpture of .. on top of a snowboard
        C) amigurumi .. wearing skis
        D) .. wearing a top hat, low poly 3D model, unreal engine
        E) brilliantly colored iridescent .. wearing a party hat
        F) DSLR full body photo of majestic .. wearing a crown, HD, green planet, national geographic 
        G) colorful 3D mosaic art suplture of .. in a Chef's Hat, unreal engine, dimensional surfaces.
        }
        From top to bottom, the animals are:
        1) elephant
        2) brown and white striped horse
        3) goldfish
        4) bee
        5) giraffe
        6) meerkat
        7) shark
    }
    \label{fig:qualitative_grid_style_unseen}
\end{figure*}

\vspace{-0.015\textheight}
\paragraph{Additional results on \animalStyle.}
We show qualitative examples of \ours on the \animalStyle dataset for both seen and unseen prompts in Fig.~\ref{fig:qualitative_grid_style_seen} and  Fig.~\ref{fig:qualitative_grid_style_unseen} respectively.
We see virtually no degradation in our performance on the unseen prompts, corroborating our quantitative results and demonstrating our model's combinatorial generalization.

\begin{figure*}%
    \includegraphics[width=\linewidth,trim={0 0 0 0},clip]{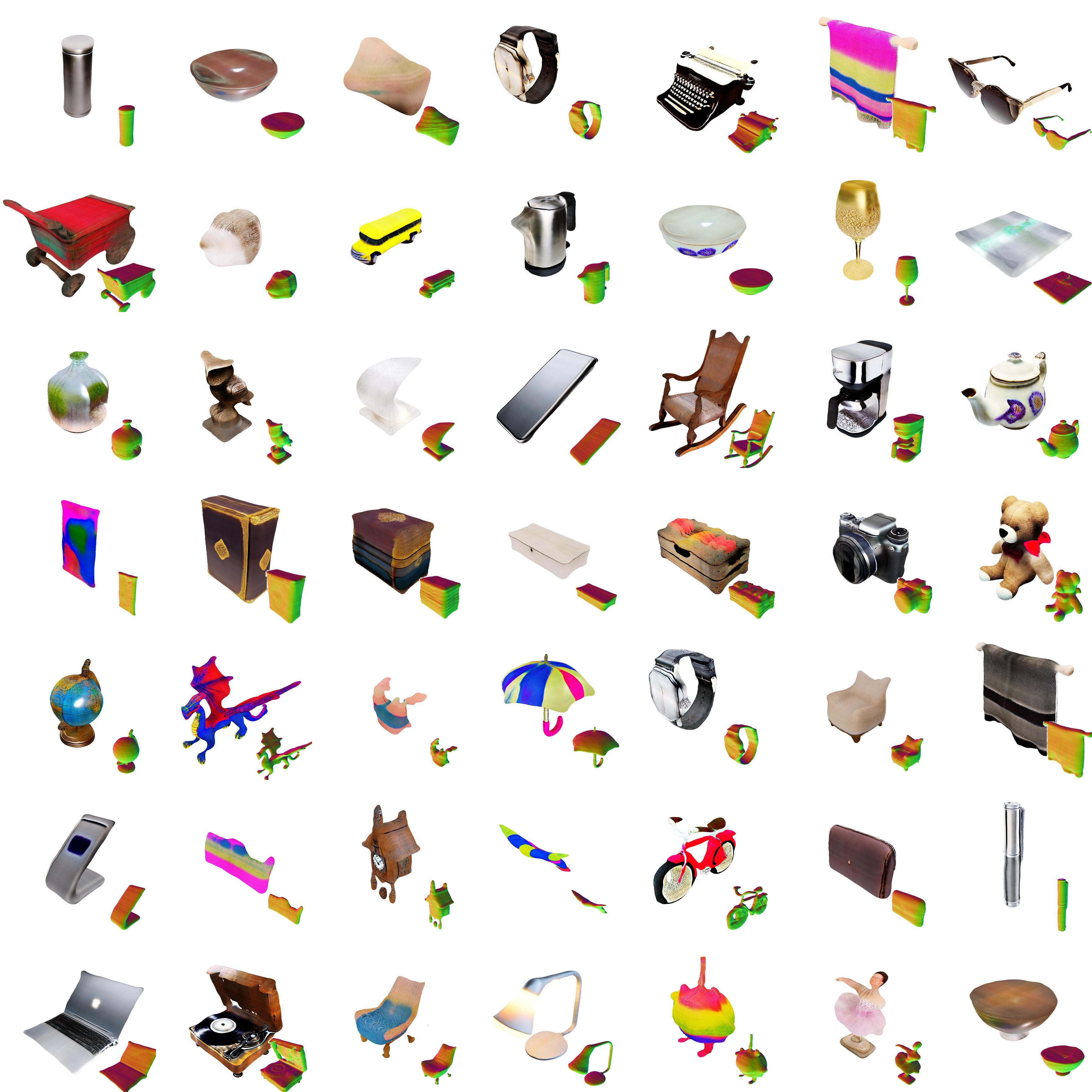}
    \caption{
        We show example results of our model on the unseen in-distribution set created using further ChatGPT suggestions. From left to right, the prompts for the first 2 rows are: 
\textit{1) A set of sleek and modern salt and pepper grinders with adjustable coarsene...; A sleek and modern salad bowl with a curved design and vibrant stripes...; A set of artisanal handmade soap bars in various shapes and colors...; A classic and timeless leather watch with a simple dial and a genuine leath...; A vintage-inspired typewriter with a metallic finish and classic keycaps...; A vibrant and patterned beach towel with a tassel fringe...; A pair of stylish sunglasses with mirrored lenses and a tortoiseshell frame...
2) A classic red wagon with sturdy wooden sides and large rubber wheels...; A charming ceramic salt and pepper shaker set shaped like playful hedgehogs...; A vibrant yellow school bus with cheerful cartoon characters painted on the...; A sleek and modern electric kettle with a transparent water level indicator...; A set of ceramic nesting bowls with hand-painted floral motifs...; A set of elegant wine glasses with a gold rim and a delicate floral pattern...; A modern glass chess set with minimalist pieces and a reflective board...}
    }
    \label{fig:qualitative_grid_lvis_unseen_id}
\end{figure*}

\vspace{-0.015\textheight}
\paragraph{Additional results on \largeDataset.}
In the main manuscript, we evaluated our model trained on \largeDataset on a set of unseen prompts from DreamFusion.
These prompts are generally further out of distribution compared to our training prompts.
To evaluate unseen prompts that are more similar to our training prompts, we use ChatGPT to suggest $50$ more prompts based on $20$ of our training prompts given as reference.
We show qualitative examples in Fig.\ref{fig:qualitative_grid_lvis_unseen_id}.

\begin{figure*}%
    \includegraphics[width=\linewidth,trim={0 0 0 0},clip]{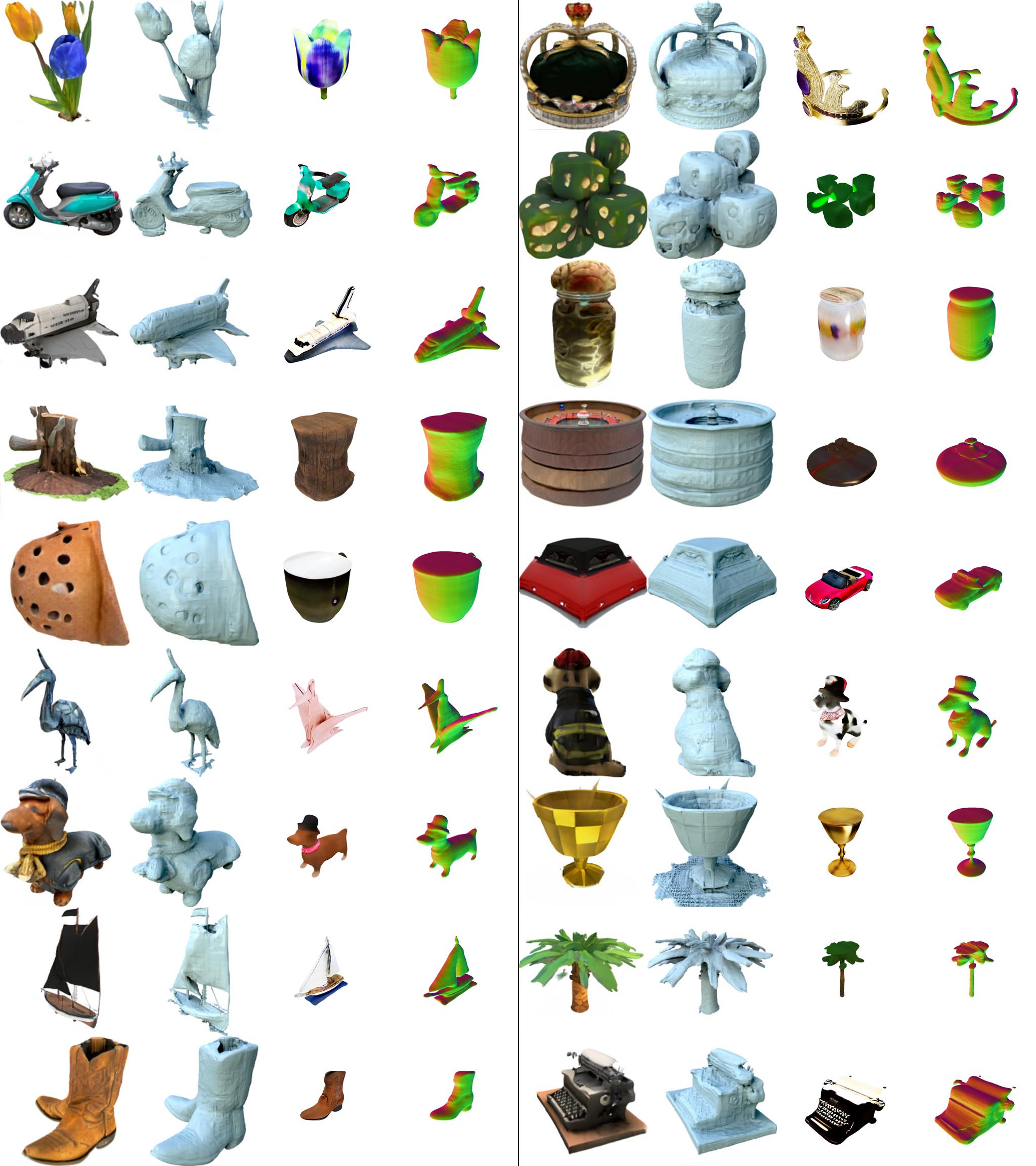}
    \vspace{-0.03\textheight}
    \caption{\textbf{Comparison between Instant3d (left) and \ours (right). }
        We display our results next to the results cropped from Instant3D paper. The Instant3D results are displayed on the left for each pair of columns, with our result on the right.
        From left to right, top to bottom the prompts are:
    \textit{a blue tulip, the Imperial State Crown of England, a teal moped, a pile of dice on a green tabletop, a Space Shuttle, a brain in a jar, a tree stump with an axe buried in it, a roulette wheel, Coffee cup with many holes, a red convertible car with the top down, an origami crane, a dalmation wearing a fireman's hat, a dachsund wearing a boater hat, a golden goblet, low poly, a majestic sailboat, a palm tree, low poly 3d model, a pair of tan cowboy boots, studio lighting, product photography, a typewriter}
    }
    \label{fig:appendix_inst3d_comp_0}
\end{figure*}

\paragraph{Additional Comparisons to Instant3D}
We apply our model trained on \largeDataset to unseen prompts from DreamFusion.
As in the main text, we select a subset of $65$ prompts closer to the training prompt distribution.
Instant3D also presents their generations for the DreamFusion prompts, so we show more qualitative side-by-side comparisons with their reported results in Fig.~\ref{fig:appendix_inst3d_comp_0}.

\paragraph{Additional Results on Test-time Optimization}
We present additional results on test-time optimization in Fig~\ref{fig:post_optimization_appendix}. Columns of Fig~\ref{fig:post_optimization_appendix} correspond to test-time optimization after $0$ (no optimization), $ 60, 120, 300$, and $600$ steps, respectively.

\subsection{Additional Ablation Studies} 

\paragraph{Additional results for ablations on stage-1.}
In the main manuscript we presented the quantitative ablations of the main components LATTE3D introduced. Due to space constraints, we only reported full quantitative metrics on seen prompts (and user preference ratings on both seen and unseen.

\begin{table}[t]
    \centering
    \caption{\small Ablation of components in stage-1 training. Trained on \largeDataset data and evaluated on unseen prompts. Preference indicate average user preference of baseline over \ours.
    }
    \label{tab:app_tab_ablate_unseen}
    \scalebox{0.8}{
    \setlength{\tabcolsep}{2pt} 
    \begin{tabular}{l|cccc|cccc}
      \toprule
       \multirow{2}{*}{Exp} & \multirow{2}{*}{MV} & \multirow{2}{*}{Unet} & \multirow{2}{*}{Pretrain} & \multirow{2}{*}{Reg} & \multicolumn{4}{c}{Unseen} \\ \cmidrule{6-9}
        & & & & & Mask-FID$\downarrow$ & Render-FID$\downarrow$ & Clip-Score$\uparrow$& Preference$\uparrow$ \\
      \midrule

       ATT3D && & & & 294.18 & 292.38 & 0.1779 & 24.2 \\
       +MV &$\checkmark$& & & & 263.70  &  256.67 & 0.1834 & 26.9 \\
       +MV+UNet &$\checkmark$& $\checkmark$ & & & 168.11 & 216.43 & 0.2283 & 45.6  \\
       +MV+UNet+PT &$\checkmark$& $\checkmark$ & $\checkmark$ &   & 176.01 & 199.09 & 0.2330 & 47.7 \\
        \hline
       \ours (S1) &$\checkmark$& $\checkmark$ & $\checkmark$ & $\checkmark$& {\bf{157.88}} & 201.17 & {\bf{0.2447}} & -\\ 
      
      \bottomrule
    \end{tabular}
    }

\end{table}

On the CLIP score, the ATT3D ablation outperforms adding MVDream.
We posit that this is due to the higher rate of Janus face artifacts when using SDS over MVDream, which can maximize the CLIP score metric at the cost of geometric errors, as evidenced by the higher Mask-FID.

\paragraph{Ablation on the upsampling layer in stage-2.} We provide an ablation study on the upsampling layer in stage-2 in Table~\ref{tab:abs_s2_upsampling} with qualitative examples in Fig.~\ref{fig:upsample_vs_noupsample_qual}. Adding the upsampling layer for the latent triplane significantly improves Render-FID and retains a strong CLIP-Score.

\begin{table}
    \centering
    \caption{
        Ablation over whether we use upsampling on the latent triplanes during stage-2 using $\blend=.9$ on the realistic animals with samples shown in Fig.~\ref{fig:upsample_vs_noupsample_qual}. Upsampling the triplane gets better performance.
    }
    \label{tab:abs_s2_upsampling}
    \begin{tabular}{l|cc}
        \toprule
        Setup & Render-FID$\downarrow$ & Clip-Score$\uparrow$ \\
        \midrule
        Stage-2 w/ upsampling & $\textbf{96.75}$ & $0.260$ \\
        Stage-2 w.o. upsampling & $104.32$& $\textbf{0.261}$ \\
        Stage-1 only & $122.24$ & $0.264$ \\
        \bottomrule 
    \end{tabular}
\end{table}
\begin{table}
    \centering
    \caption{Ablation over different guidance at stage-2. Depth-SD and SD are significantly better than MVDream and VSD. While Depth-SD and SD are close to each other regarding the render-FID and CLIP score, our user study shows that Depth-SD is still preferable. The reason is that the texture generated with Depth-SD is more faithful to the input shape.}
    \begin{tabular}{l|ccc}
        \toprule
        Guidance & Render-FID$\downarrow$ & CLIP-Score$\uparrow$ & User-Preference ($\%
        $)$\uparrow$\\
        \midrule
        Depth-SD & $96.75$ & $0.260$ & \textbf{30.56}\\
        SD & $\textbf{94.42}$ & $\textbf{0.266}$ & 29.11\\
        MVDream & $110.13$ & $0.248$ & 18.33\\
        VSD & $102.92$ & $0.250$ & 22.00\\
        \bottomrule 
    \end{tabular}
    \label{tab:abs_s2_guidance}
\end{table}

\paragraph{Ablation on stage-1 guidance choices.}
We contrast the SDS guidance in stage-1 between Stable Diffusion~\cite{Rombach_2022_CVPR} and MVDream~\cite{shi2023MVDream}. We provide qualitative examples in Figs.~\ref{fig:stage1_vs_stage_2_qual} and~\ref{fig:blend_scale_comparison}.
Notably, Fig.~\ref{fig:stage1_vs_stage_2_qual} shows examples of prompts where stage-1 SDS can produce incorrect geometry leading to Janus-face-type artifacts propagating to the second stage, while stage-1 MVDream avoids Janus-faces.

\input{images/s1_ablate_fig}

As MVDream is already finetuned on 3D consistent data, its geometric consistency is already quite good.
In the first stage, reference geometry regularization is especially important when using Stable Diffusion instead of MVDream.

\begin{figure}
    \vspace{-0.015\textheight}
    \centering
    \includegraphics[width=0.9\linewidth]{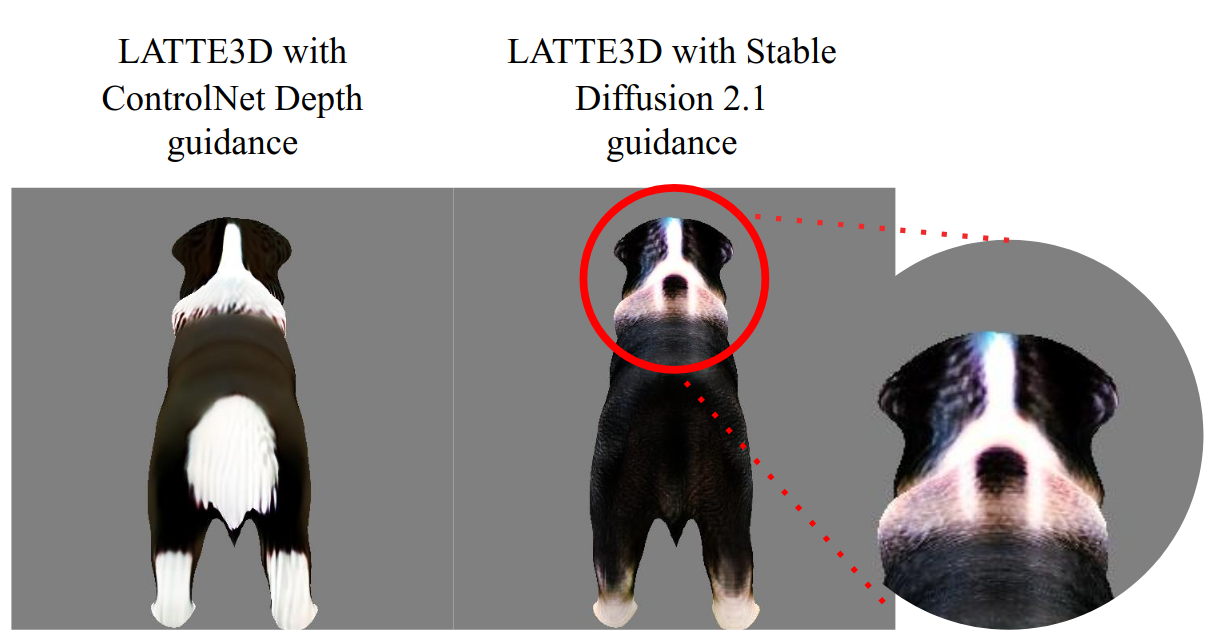}
    \vspace{-0.015\textheight}
    \caption{We compare training \ours stage2 with depth-conditioned ControlNet against Stable Diffusion. Without depth conditioning, we observe the Janus face issue~\cite{poole2022dreamfusion} in textures.}
    \label{fig:s2_ablate}
    \vspace{-0.015\textheight}
\end{figure}

\paragraph{Ablation on stage-2 guidance choices.}
In Tab.~\ref{tab:abs_s2_guidance}, we explore different guidance choices in our stage-2 training. Overall, Depth-SD and SD perform consistently better than MVDream and VSD. MVDream performs poorly in the refinement stage because it inputs lower-resolution images. We found that SD and Depth-SD perform close to each other quantitatively. We use Depth-SD since it occasionally resolves Janus face issues (Fig.~\ref{fig:s2_ablate}).

\section{Failure Cases and Limitations}
\label{sec:app_failure}

\begin{figure}[h]%
    \centering
    \includegraphics[width=0.8\linewidth,trim={0 4cm 4cm 0},clip]{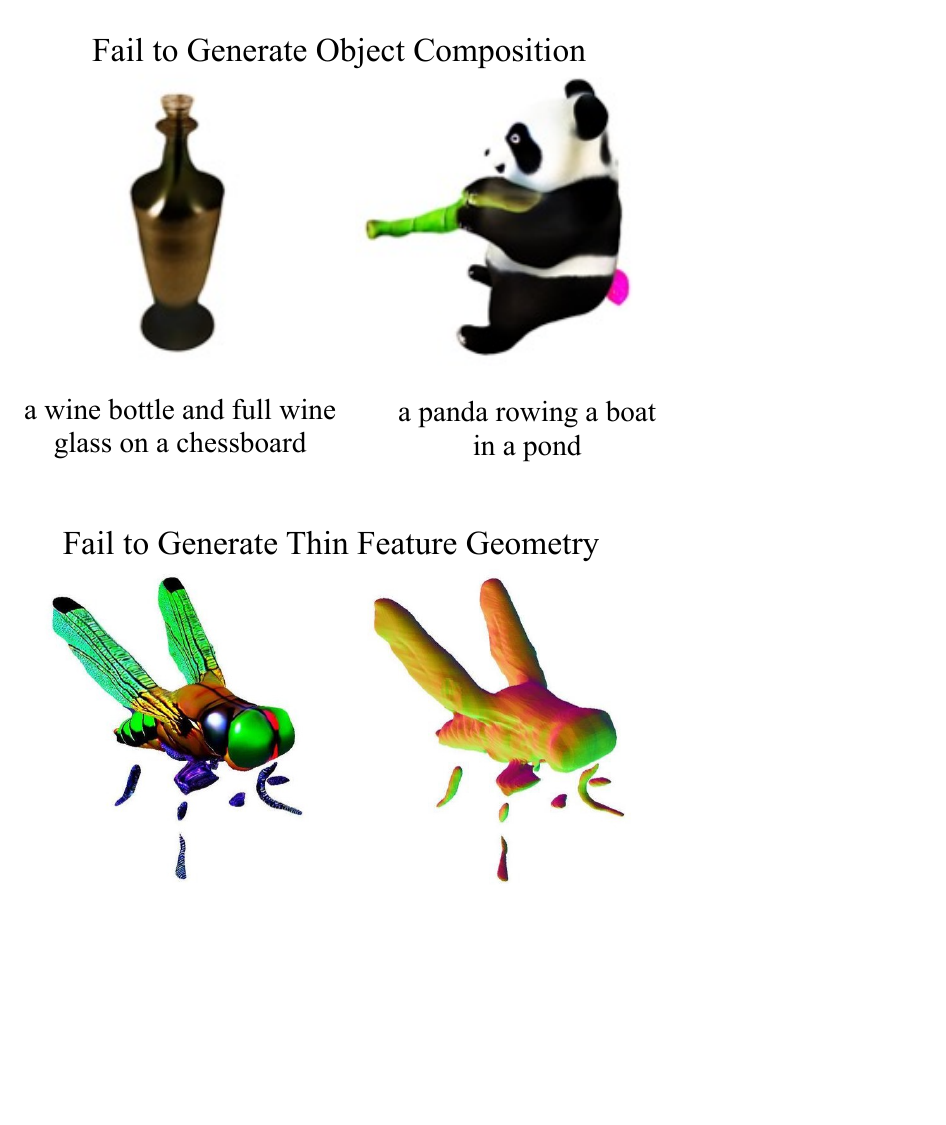}
    \caption{
    We show our model's failure to generate compositions of objects following the prompt (top row) and thin feature geometry details (bottom row).
    }
    \label{fig:appendix_limitation_examples}
\end{figure}

We display typical failure cases of our model in Fig.~\ref{fig:appendix_limitation_examples}.
Our model trained on \largeDataset can miss details in composed prompts and often only generates one object in multi-object prompts.
We expect such failure cases to diminish by increasing the diversity of prompts used for training.
Some thin feature details may also be lost since geometry is frozen from stage-1 and converted from volume to surface rendering. Stabilizing geometry training in stage-2 would help recover some of these details.

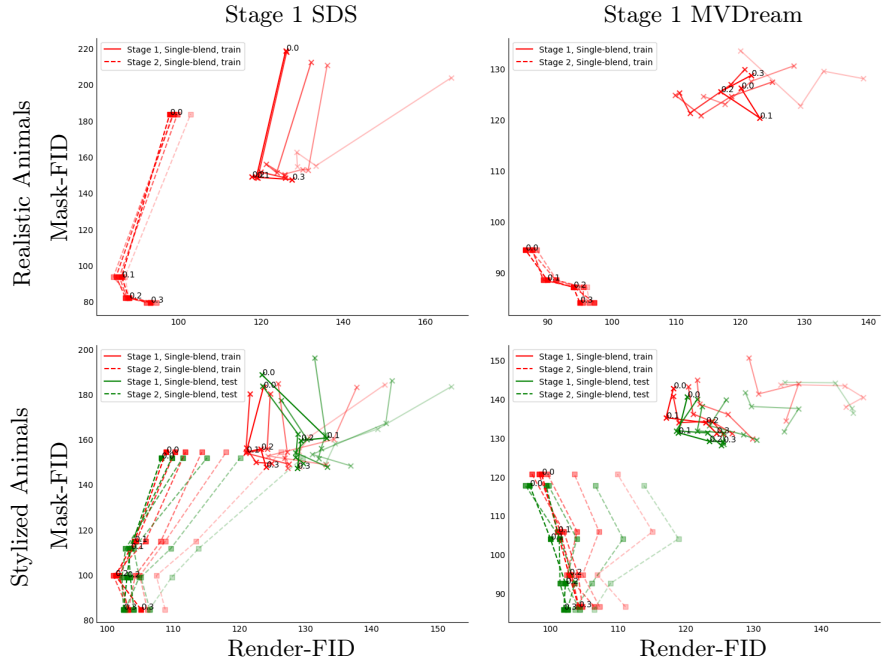
\begin{figure*}%
    \centering
    \begin{tikzpicture}
        \centering
        \node (img11){\includegraphics[trim={0.9cm 1.0cm 0.2cm 1.0cm}, clip, width=.43\linewidth]{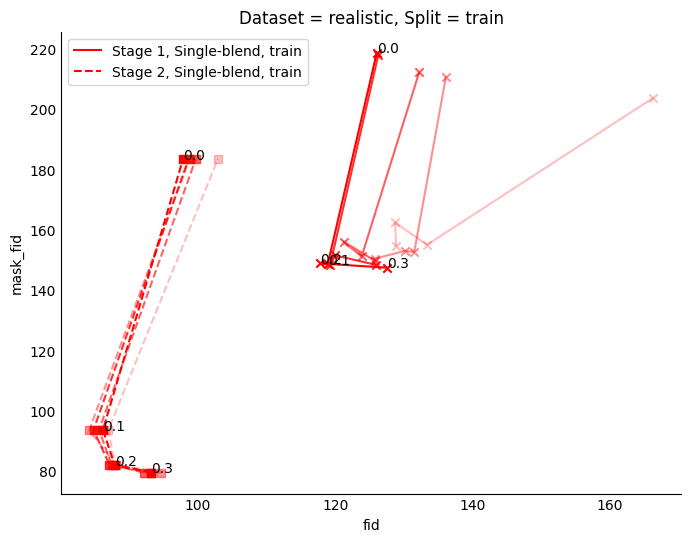}};
        \node [right=of img11, xshift=-1cm](img12){\includegraphics[trim={0.9cm 1.0cm 0.2cm 1.0cm}, clip, width=.43\linewidth]{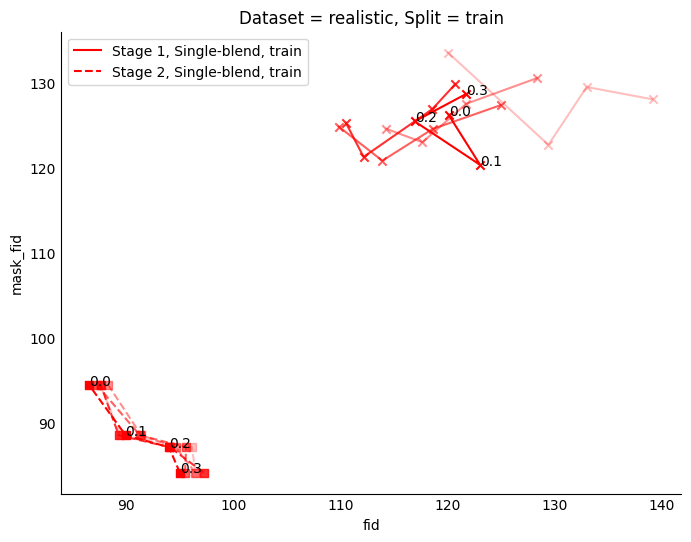}};

        \node [below=of img11, yshift=1.0cm](img21){\includegraphics[trim={0.9cm 1.0cm 0.2cm 1.0cm}, clip, width=.43\linewidth]{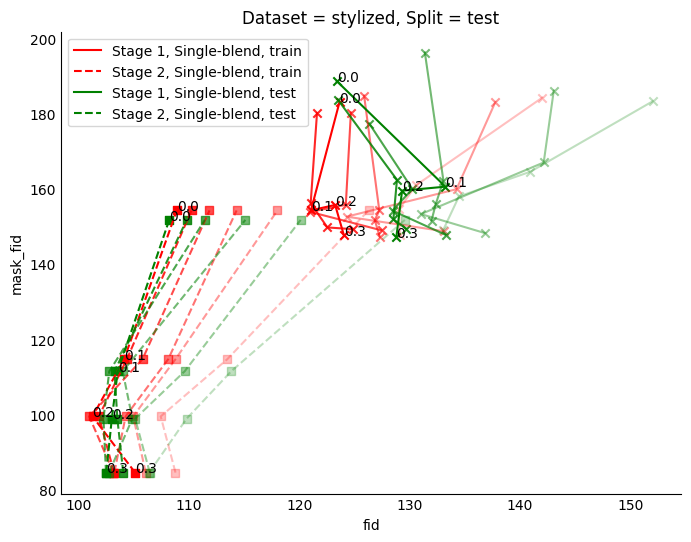}};
        \node [right=of img21, xshift=-1cm](img22){\includegraphics[trim={0.9cm 1.0cm 0.2cm 1.0cm}, clip, width=.43\linewidth]{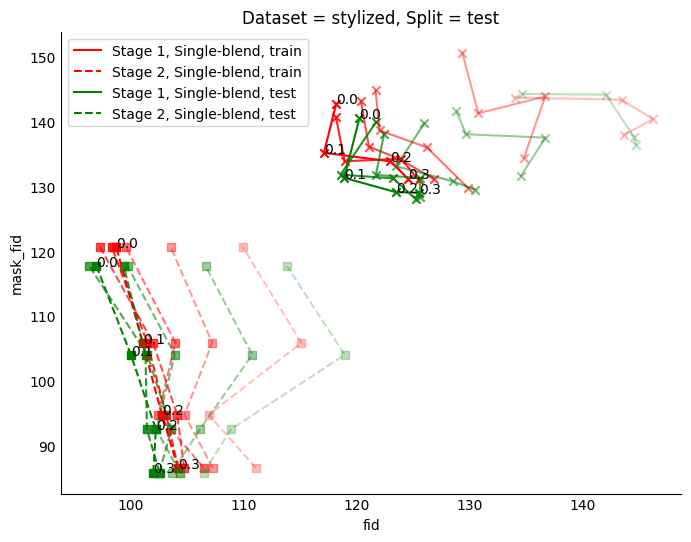}};

        \node[left=of img11, node distance=0cm, rotate=90, xshift=0.75cm, yshift=-.9cm,  font=\color{black}]{Mask-FID};
        \node[left=of img21, node distance=0cm, rotate=90, xshift=0.75cm, yshift=-.9cm,  font=\color{black}]{Mask-FID};
        \node[below=of img21, node distance=0cm, xshift=0.25cm, yshift=1.15cm,  font=\color{black}]{Render-FID};
        \node[below=of img22, node distance=0cm, xshift=0.25cm, yshift=1.15cm,  font=\color{black}]{Render-FID};
        \node[left=of img11, node distance=0cm, rotate=90, xshift=1.2cm, yshift=-.4cm,  font=\color{black}]{Realistic Animals};
        \node[left=of img21, node distance=0cm, rotate=90, xshift=1.2cm, yshift=-.4cm,  font=\color{black}]{Stylized Animals};
        \node[above=of img11, node distance=0cm, xshift=0.25cm, yshift=-1.1cm,  font=\color{black}]{Stage 1 SDS};
        \node[above=of img12, node distance=0cm, xshift=0.25cm, yshift=-1.1cm,  font=\color{black}]{Stage 1 MVDream};
    \end{tikzpicture}
    \caption{
        We show the frontier of the tradeoff between Mask FID and Render-FID at various blending $\blend$ (annotated) on the realistic (top) and stylized (bottom) animals for training (red) and testing (green) prompts in stage-1 (solid lines) and stage-2 (dashed lines) as optimization progresses from low-alpha in the start to high-alpha at the end of training.
        We display results of $5$ evenly spaced points in training for stages 1 and 2, with optimization horizons of $50$k iterations for realistic and $100$k for stylized.
        Notably, the gap between seen (training) and unseen (testing) prompts at the end of training is small, showing effective generalization.
        Our Render-FID improves over training with a large jump from stage-1 to 2.
        Our Mask-FID improves at the start of stage-2 when we convert to surface rendering.
        The mask-FID is often lower for larger shape regularization, showing shape regularization improves consistency with the source object, illustrating that we can use $\blend$ to enhance user-controllability towards the input.
    }
    \label{fig:main_ablation_quant}
\end{figure*}

\input{images/upsample_vs_noupsample_ablate}

\clearpage
\begin{figure*}%
    \includegraphics[width=\linewidth,trim={0 0 0 0},clip]{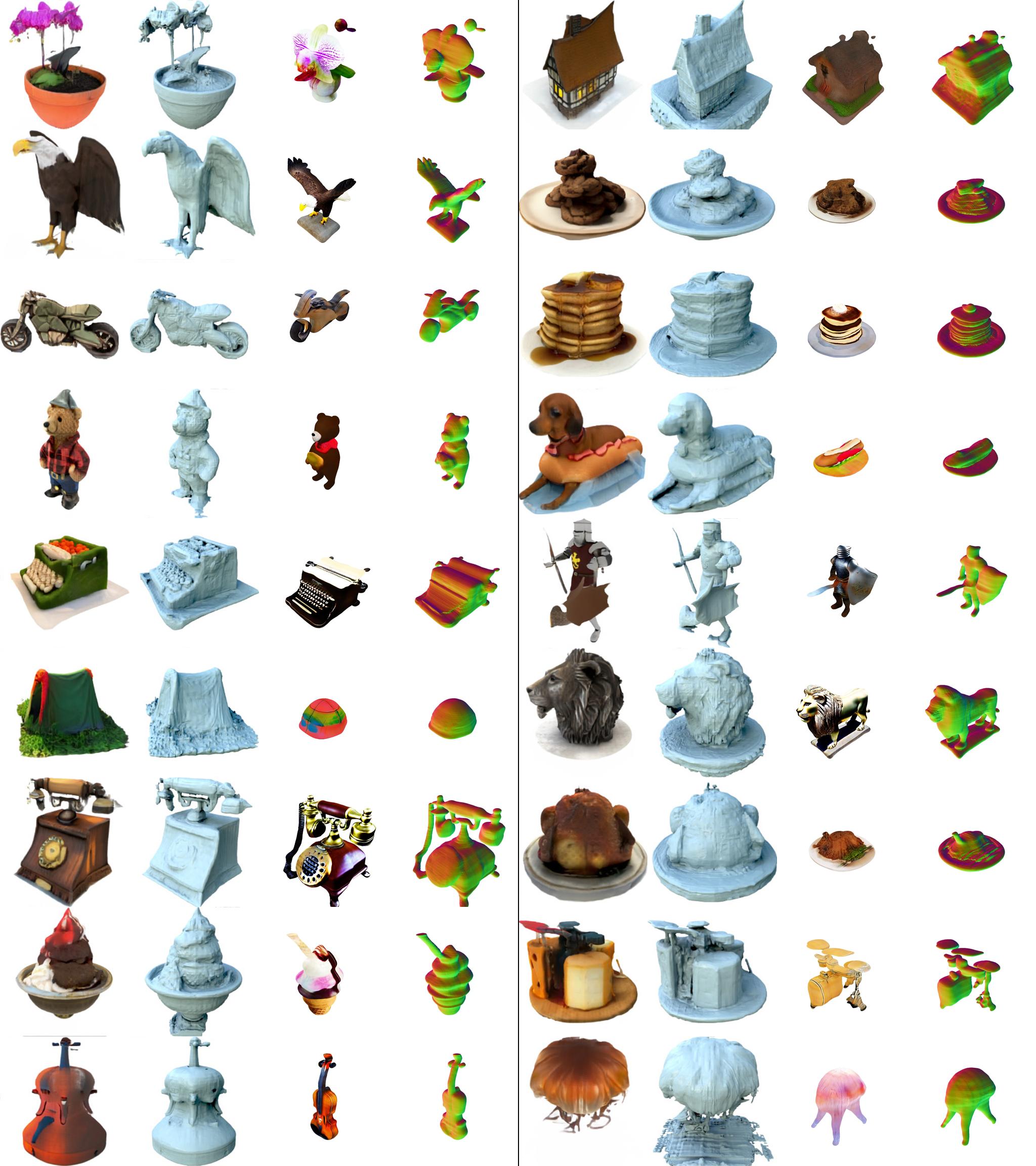}
    \vspace{-0.03\textheight}
    \caption{ \textbf{Comparison between Instant3d (left) and \ours (right). }
        From left to right, top to bottom, the prompts are:
    \textit{an orchid flower planted in a clay pot, a model of a house in Tudor style, a bald eagle, a plate piled high with chocolate chip cookies, an origami motorcycle, a stack of pancakes covered in maple syrup, a bear dressed as a lumberjack, a hotdog in a tutu skirt, edible typewriter made out of vegetables, a knight chopping wood, a colorful camping tent in a patch of grass, a metal sculpture of a lion's head, highly detailed, a rotary telephone carved out of wood, a roast turkey on a platter, an ice cream sundae, a drum set made of cheese, a beautiful violin sitting flat on a table, a lion's mane jellyfish}
    }
    \label{fig:appendix_inst3d_comp_1}
\end{figure*}
\clearpage
\begin{figure*}%
    \includegraphics[width=\linewidth,trim={0 0 0 0},clip]{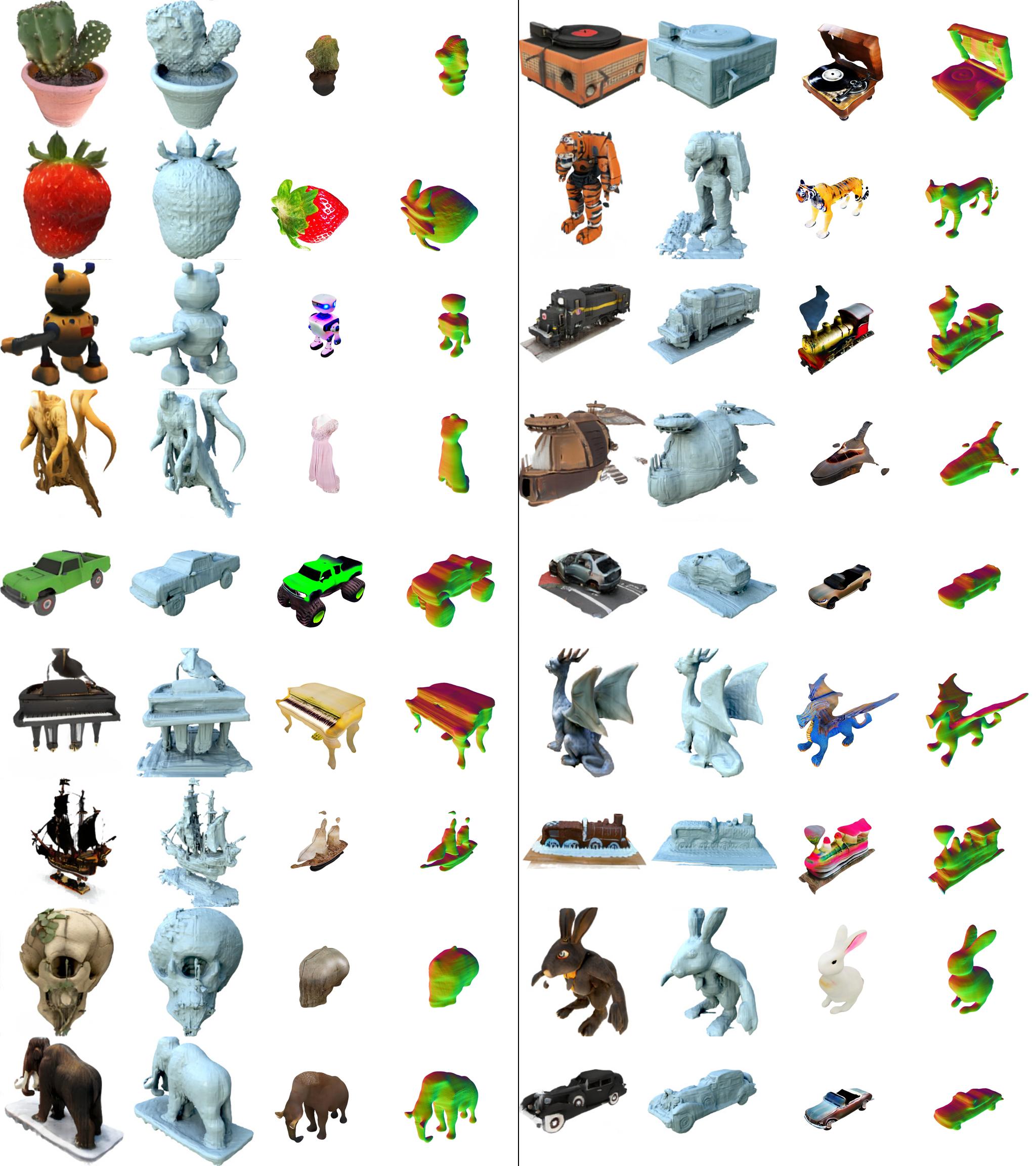}
    \vspace{-0.03\textheight}
    \caption{ \textbf{Comparison between Instant3d (left) and \ours (right). }
        From left to right, top to bottom, the prompts are:
    \textit{a small saguaro cactus planted in a clay pot, a vintage record player, a ripe strawberry, a robot tiger, a toy robot, Two locomotives playing tug of war, Wedding dress made of tentacles, a steampunk space ship designed in the 18th century, a green monster truck, a completely destroyed car, a baby grand piano viewed from far away, a porcelain dragon, a spanish galleon sailing on the open sea, a cake in the shape of a train, a human skull with a vine growing through one of the eye sockets, a rabbit, animated movie character, high detail 3d model, a woolly mammoth standing on ice, a classic Packard car}
    }
    \label{fig:appendix_inst3d_comp_2}
\end{figure*}
\clearpage
\begin{figure*}%
    \includegraphics[width=\linewidth,trim={0 0 0 0},clip]{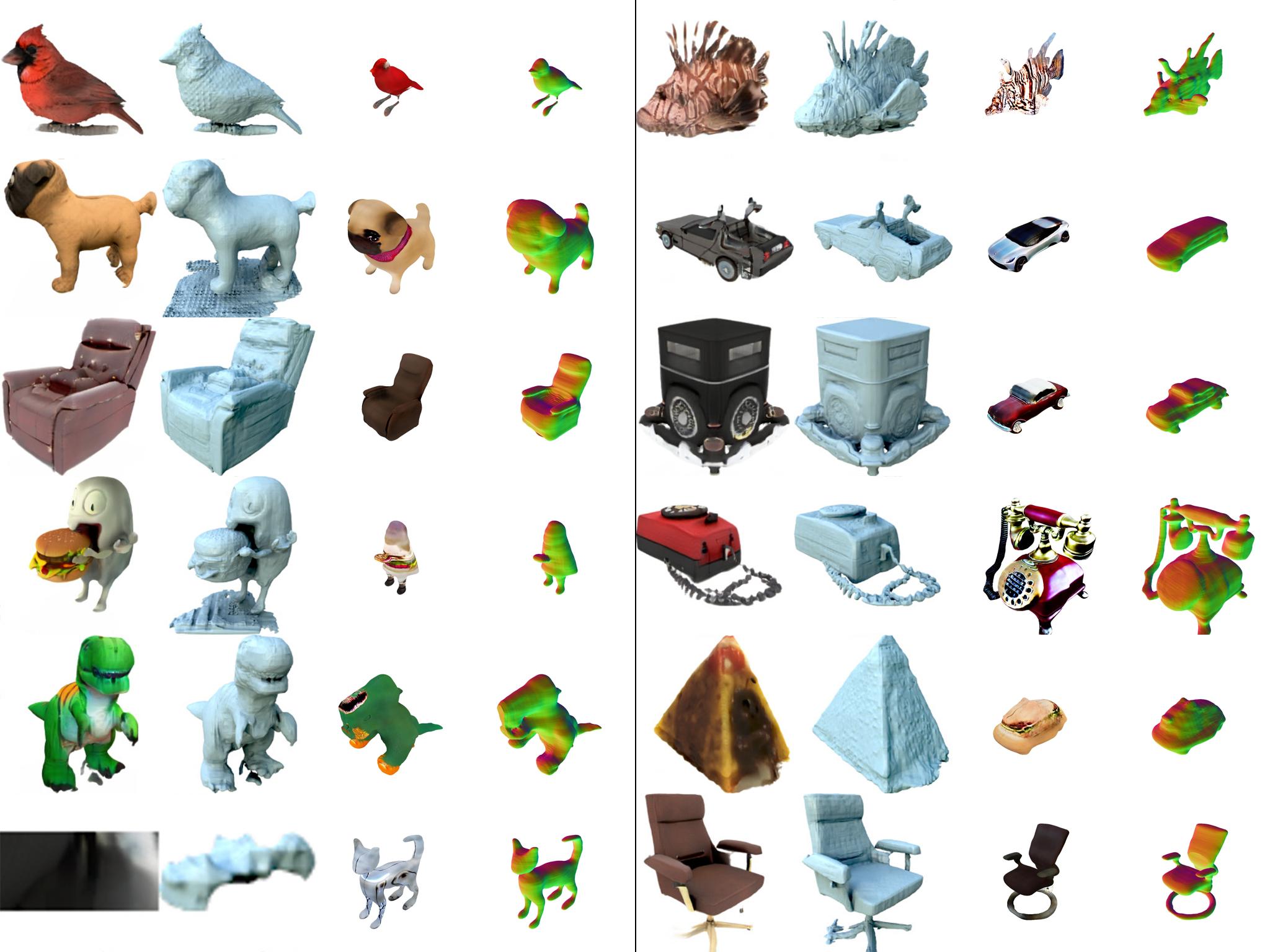}
    \vspace{-0.03\textheight}
    \caption{ \textbf{Comparison between Instant3d (left) and \ours (right). }
        From left to right, top to bottom, the prompts are:
   \textit{ a red cardinal bird singing, a lionfish, a pug made out of modeling clay, A DMC Delorean car, a recliner chair, an old vintage car, a ghost eating a hamburger, a red rotary telephone, a plush t-rex dinosaur toy, studio lighting, high resolution, A DSLR photo of pyramid shaped burrito with a slice cut out of it, a shiny silver robot cat, an expensive office chair}
    }
    \label{fig:appendix_inst3d_comp_3}
\end{figure*}
\clearpage

\input{images/blend_scale_comparison}

\begin{figure*}
    \centering
        \setlength{\tabcolsep}{0.1pt} 
    \begin{subfigure}{\linewidth}
  \centering
  \begin{tabular}{ccccc} 
    \hspace*{9mm} 400ms \hspace*{7.5mm} & \hspace*{7.5mm} 1min \hspace*{7.5mm} & \hspace*{7.5mm} 2min \hspace*{7.5mm} & \hspace*{7.5mm} 5min \hspace*{7.5mm} & \hspace*{7.5mm} 10min \hspace*{9mm} \\
    \multicolumn{5}{c}{\includegraphics[width=\linewidth]{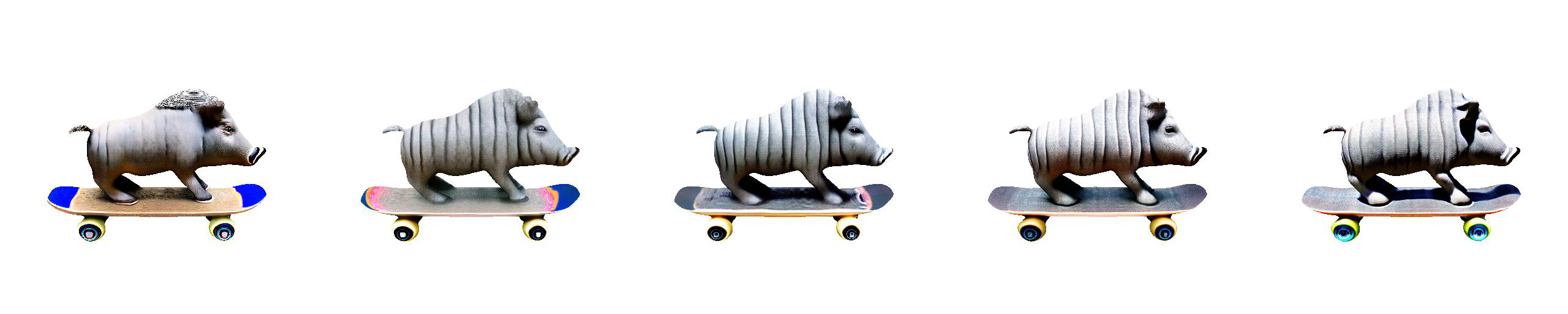}} \\
    \multicolumn{5}{c}{\includegraphics[width=\linewidth]{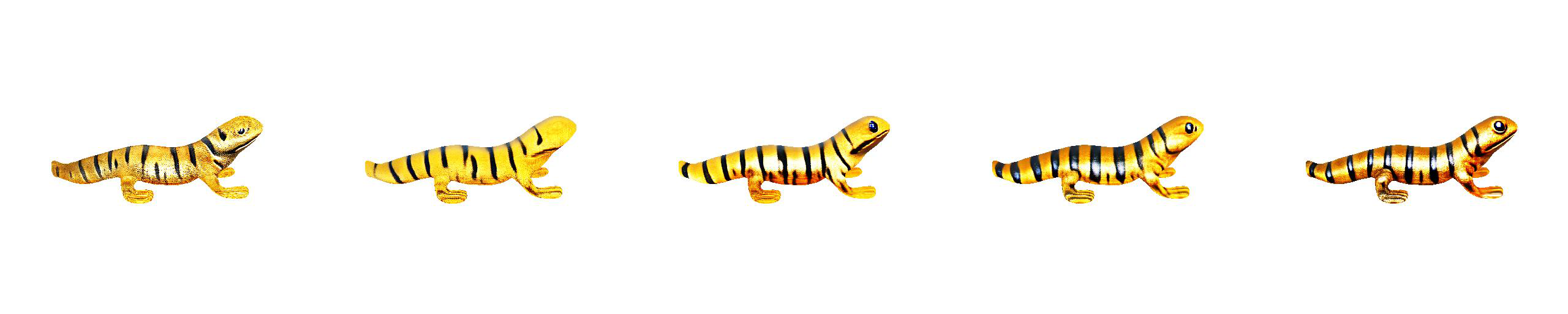}} \\
    \multicolumn{5}{c}{\includegraphics[width=\linewidth]{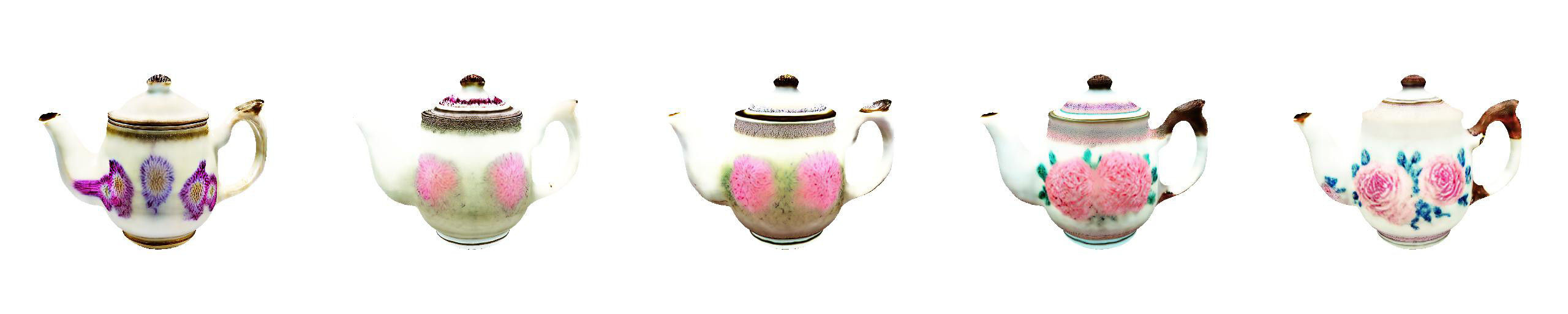}} \\
    \multicolumn{5}{c}{\includegraphics[width=\linewidth]{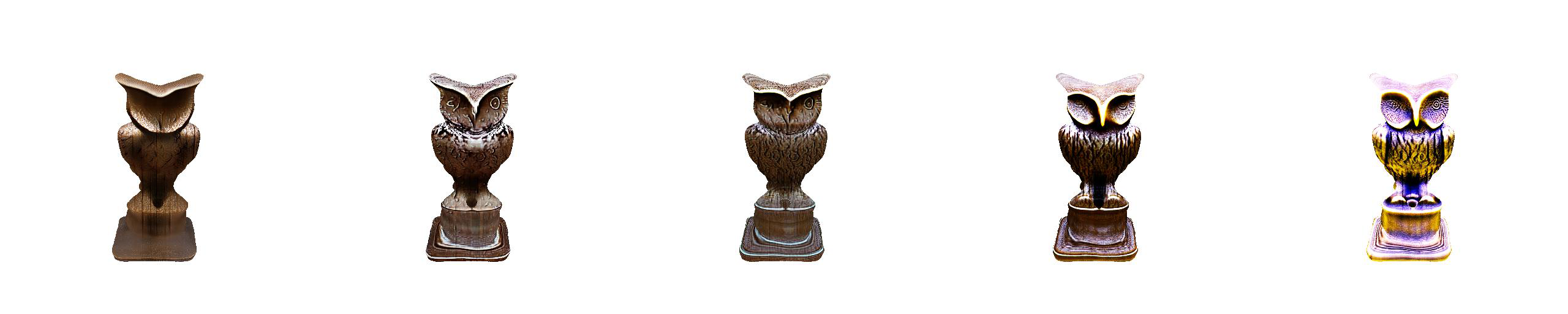}} \\
    \multicolumn{5}{c}{\includegraphics[width=\linewidth]{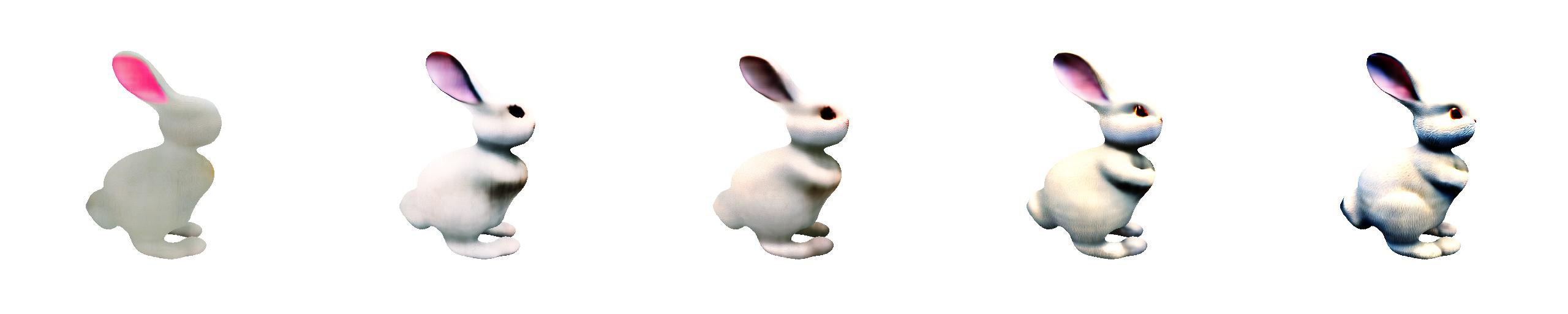}} \\
    \multicolumn{5}{c}{\includegraphics[width=\linewidth]{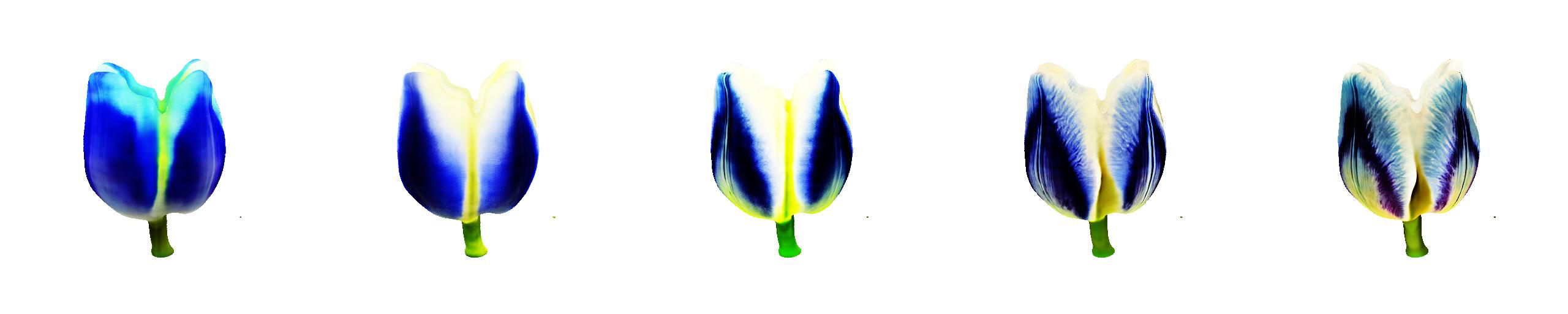}}

  \end{tabular}
\end{subfigure}

    \caption{Test time optimization. First column: without test time optimization, $\generateTime$. Other columns: test time optimization with the given time budget denoted. 
    }
    \label{fig:post_optimization_appendix}
\end{figure*}

\newpage
\section{Additional New Capability: User-controllability via Amortizing the Regularization Strength}\label{sec:app_amort_reg}
We also train our model in a way that allows users to select the regularization strength $\blend$ at inference.
This provides another axis for users to guide generation with 3D objects in real-time, analogous to guiding generated results towards provided images in text-to-image generation tools.

We do this by simultaneously training on multiple different $\blend$ (uniformly sampled in a range), which are also input to the network -- i.e., amortizing over $\blend$.
We input $\blend$ simply by linearly scaling the text-conditional cross-attention residual added at each layer by $1 - \blend$.
Intuitively, this forces the magnitude of the text influence on the features to be smaller for higher $\blend$ and prevents any dependence at $\blend=1$.

We contrast results with and without amortized loss weighting qualitatively and quantitatively in Figs.~\ref{fig:amort_vs_nonamort_qual} and \ref{fig:amort_vs_nonamort_quant} respectively, showing a minimal loss in quality from non-amortized runs.
This means that, for a small impact on quality and compute, we can allow a user to select the loss weighting dynamically at inference time.

Fig.~\ref{fig:interpolation_pc_text} illustrates an example where we blend the RGB loss from the 3D object, allowing users to smoothly interpolate between the 3D object and the object from only the text-prompt.
This is a 3D analog to the feature of text-to-image models having a test-time user-controlled weighting towards a source image.
The interpolation is a 4D (a 3D animation), generating a separate 3D object for each $\blend \in [0, 1]$.

\input{images/blendamort_vs_single}

\section*{Author Contributions}
\textbf{All authors} have significant contributions to ideas, explorations, and paper writing.
Specifically, \textbf{XHZ} and \textbf{GJ} led the research and organized team efforts.
\textbf{KX} developed fundamental code for experiments and led the experiments of pre-training, stage-1 training, data curation, and evaluation.
\textbf{JoL} led experiments for stage-2 training and evaluation. 
\textbf{TSC} led experiments for integrating MVDream prior, stage-1 training, and test-time optimization.
\textbf{JaL} led experiments for baselines for user study.
\textbf{GJ} led the data preprocessing, pre-training, and user study. 
\textbf{XHZ} led experiments for stage-1 training and data curation. 
\textbf{SF} advised the research direction and designed the scope of the project.

\end{document}